%% file: 0Abstract.tex
\documentclass{article}

 \usepackage[preprint]{neurips_2026}

\usepackage[utf8]{inputenc} %
\usepackage[T1]{fontenc}    %
\usepackage{url}            %
\usepackage{booktabs}       %
\usepackage{amsfonts}       %
\usepackage{nicefrac}       %
\usepackage{microtype}      %

\usepackage{etoc}
\usepackage{enumitem}
\usepackage{graphicx}
\usepackage{amsmath}
\usepackage[normalem]{ulem}
\usepackage{multirow}
\usepackage{caption}
\usepackage{wrapfig}
\usepackage{stmaryrd}
\usepackage[table]{xcolor}
\usepackage[most]{tcolorbox}
\usepackage[colorlinks, linkcolor=mydarkred, citecolor=myblue]{hyperref}
\definecolor{mydarkblue}{rgb}{0,0.08,0.45}
\definecolor{mydarkred}{rgb}{0.6,0,0}
\definecolor{myblue}{HTML}{268BD2}
\definecolor{mygreen}{HTML}{658354}
\definecolor{results}{RGB}{220, 230, 240}

\title{GeoRouter: Dynamic Paradigm Routing for Worldwide Image Geolocalization}

\author{
    Pengyue Jia$^{1}$, Derong Xu$^{1}$, Yingyi Zhang$^1$, Xiaopeng Li$^1$, Wenlin Zhang$^1$\\
    \textbf{Yi Wen}$^1$, \textbf{Yuanshao Zhu}$^1$, \textbf{Xiangyu Zhao}$^1$ \\
    $^1$Department of Data Science, City University of Hong Kong,   \\
    \texttt{jia.pengyue@my.cityu.edu.hk,xianzhao@cityu.edu.hk }
}

\begin{document}
\maketitle
\begin{abstract}
  Worldwide image geolocalization aims to predict precise GPS coordinates for images captured anywhere on Earth, which is challenging due to the large visual and geographic diversity. Recent methods mainly follow two paradigms: retrieval-based approaches that match queries against a reference database, and generation-based approaches that directly predict coordinates using Large Vision-Language Models (LVLMs). 
  However, we observe distinct error profiles between them: retrieval excels at fine-grained instance matching, while generation offers robust semantic reasoning. This complementary heterogeneity suggests that no single paradigm is universally superior.
  To harness this potential, we propose \textbf{GeoRouter}, a dynamic routing framework that adaptively assigns each query to the optimal paradigm. GeoRouter leverages an LVLM backbone to analyze visual content and provide routing decisions. To optimize GeoRouter, we introduce a distance-aware preference objective that converts the distance gap between paradigms into a continuous supervision signal, explicitly reflecting relative performance differences. Furthermore, we construct \textbf{GeoRouting}, the first large-scale dataset tailored for training routing policies with independent paradigm predictions. Extensive experiments on IM2GPS3k and YFCC4k demonstrate that GeoRouter significantly outperforms state-of-the-art baselines. 
\end{abstract}
\input{1Introduction}
\input{2RelatedWork}
\input{3Methodology}
\input{4Experiments}
\input{5Conclusion}

\newpage
\bibliography{references}{}
\bibliographystyle{unsrt}

\clearpage
\appendix
\begin{center}
    \LARGE \textbf{Technical Appendix}
    \vspace{1em}
\end{center}
\addcontentsline{toc}{part}{Appendix}

\begingroup
  \etocsetnexttocdepth{subsection}
  \etocsettocstyle{\section*{Table of Contents}}{}
  \localtableofcontents
\endgroup
\clearpage

\input{6Appendix}

\end{document}

%% file: 1Introduction.tex
\section{Introduction} \label{sec:intro}

Worldwide image geolocalization~\cite{wilson2021visual,li2025pixels} aims to determine the precise geographic coordinates of an image captured anywhere on Earth. In contrast to restricted localization tasks that limit the space to specific cities or known landmarks~\cite{noh2017large,cao2020unifying,tan2021instance,lee2022correlation,shao2023global}, this global scale approach supports a broader range of open-world applications, 
including crime tracking~\cite{liu2024image,li2025cross,kadha2025unravelling}, environmental monitoring~\cite{wang2024llmgeo}, and autonomous navigation~\cite{lin2013cross}. 
However, achieving high global accuracy remains challenging due to immense environmental diversity. Models must distinguish visually similar but geographically distant scenes across varied climates, vegetation, and cultures~\cite{astruc2024openstreetview,jay2025evaluating,pandegeochain,talreja2026georc}.

\begin{figure}
    \centering
    \includegraphics[width=\linewidth]{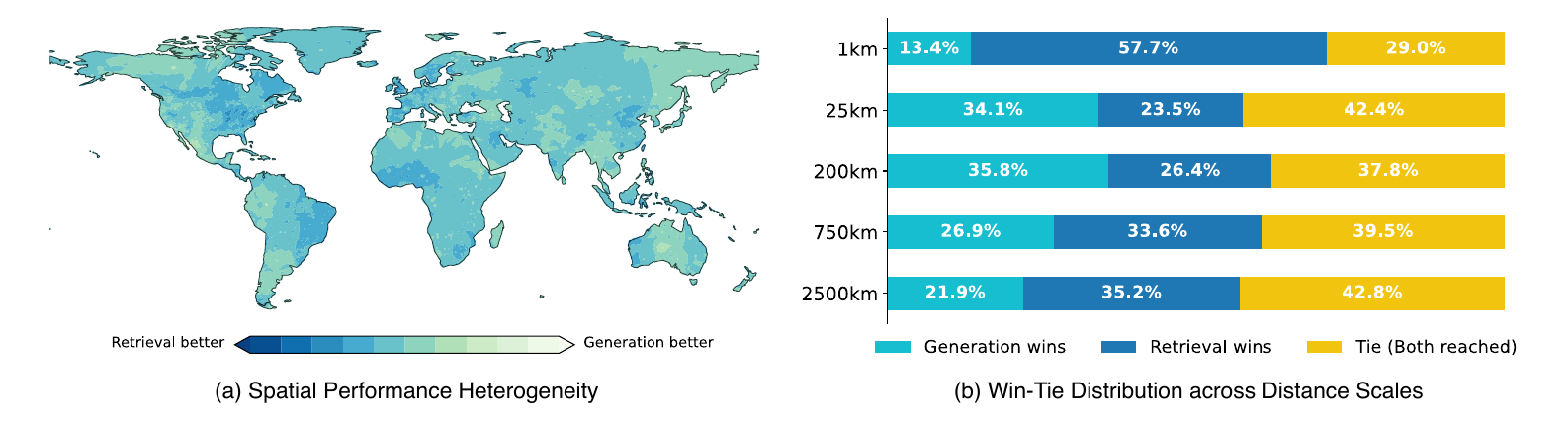}
    \caption{Paradigm Complementarity in Worldwide Image Geolocalization.}
    \label{fig:intro}
\end{figure}

Recent state-of-the-art methods generally adhere to two paradigms: (1)  retrieval-based methods~\cite{haas2023learning,vivanco2023geoclip,jia2025georanker}, which predict coordinates by matching query images against a reference database, and (2) generation-based methods~\cite{fang2026geomr,zhou2024img2loc,jia2024g3,ghasemi2025geotoken}, which utilize Large Vision-Language Models (LVLMs) to directly predict locations~\cite{wu2026vision,yu2026locatability}, invoke external tools~\cite{ji2026thinking,zheng2026learningwanderimprovingglobal}, or apply semantic reasoning for spatial inference~\cite{li2024georeasoner,li2025recognition,wang2025gre,jin2026geoagentlearninggeolocatereinforced,shi2026geobayes,su2026interpretable}. To systematically compare the capabilities of these paradigms, we conduct an empirical study on the MP16 dataset~\cite{DBLP:conf/mediaeval/ChoiHLT16}, deploying GeoRanker~\cite{jia2025georanker} and Gemini-2.5-pro~\cite{comanici2025gemini} as representative models for retrieval and generation, respectively. As illustrated in Figure~\ref{fig:intro}(a), our analysis reveals a substantial spatial performance heterogeneity, where the optimal method varies distinctly across different geographic regions rather than one model dominating globally.
Furthermore, the quantitative breakdown in Figure~\ref{fig:intro}(b) reveals two key patterns. First, retrieval-based methods achieve higher success rates at fine-grained (1km) and coarse (750km, 2500km) thresholds, while generation-based methods perform better at intermediate levels (25km, 200km). Second, at any given threshold, both paradigms each win on a substantial fraction of queries, confirming that neither method consistently dominates. These observations indicate that the two paradigms exhibit complementary error profiles rather than simply overlapping in capability.
Despite this, existing research~\cite{dufour2025around,wanglocdiff,haas2024pigeon,dou2024gaga} primarily focuses on the isolated optimization of individual paradigms, overlooking the potential for holistic integration. This observation motivates a critical research question: \textit{How can we dynamically route each query to the most suitable paradigm to break the performance ceiling of single-model systems?}

To address this challenge, we introduce GeoRouter, a unified framework designed to tackle the novel task of geolocalization routing. Specifically, GeoRouter leverages an LVLM as its backbone to exploit advanced visual understanding and spatial reasoning capabilities. By analyzing the semantic content of the given contextual information, the model predicts a scalar routing score that adaptively assigns the query to either the retrieval-based or the generation-based paradigm. To robustly optimize this decision mechanism, we propose the Distance-Aware Preference Optimization (DisPO) objective. Unlike standard binary classification, which treats routing as a discrete choice, DisPO transforms the geodesic distance disparity between paradigm predictions into a continuous supervision signal. This enables the model to discern not only which paradigm provides a more accurate prediction but also the magnitude of the performance gain, ensuring precise and calibrated routing decisions.

To facilitate the training of GeoRouter, we construct GeoRouting, the first large-scale dataset designed for the geolocalization routing task. Unlike standard geolocalization datasets that only provide ground-truth coordinates, GeoRouting includes performance comparisons between paradigms.
This rich supervision allows the model to learn the comparative advantage of each paradigm across diverse queries, rather than focusing on direct coordinate prediction.
Consequently, GeoRouting can serve as a foundational benchmark to standardize and advance the study of dynamic model selection in image worldwide geolocalization.
We conduct extensive experiments on two well-established benchmarks, IM2GPS3K~\cite{hays2008im2gps} and YFCC4K~\cite{thomee2016yfcc100m}, to validate the effectiveness of our approach. The results demonstrate that GeoRouter achieves high routing accuracy and substantially improves overall geolocalization performance. Furthermore, we perform comprehensive ablation studies and hyperparameter analyses to verify the necessity of each component and to assess the impact of different configurations on the final results. Our contributions can be summarized as follows:
\begin{enumerate}[leftmargin=*]
    \item We systematically reveal the performance complementarity between retrieval and generation paradigms and formulate the novel routing task to break single-model performance ceilings.
    \item We propose GeoRouter, a VLM-based framework utilizing DisPO to enable continuous confidence modeling for robust and precise dynamic routing.
    \item We release GeoRouting, the first large-scale routing dataset for worldwide image geolocalization, establishing a standard for this emerging research direction.
    \item Extensive experiments on IM2GPS3K and YFCC4K demonstrate that GeoRouter achieves comprehensive improvements over baselines, proving the efficacy of the dynamic routing paradigm.
\end{enumerate}

%% file: 2RelatedWork.tex
\section{Related Work}

\subsection{Image Geolocalization}

Image geolocalization~\cite{sarkar2024gomaa,vo2017revisiting} aims to predict the precise geographic coordinates of an image captured anywhere on Earth, and it is an important interdisciplinary task at the intersection of computer vision~\cite{voulodimos2018deep,he2016deep} and geographic artificial intelligence~\cite{mai2024opportunities,janowicz2020geoai}. Early approaches~\cite{weyand2016planet,seo2018cplanet,muller2018geolocation,pramanick2022world,clark2023we} formulated this problem as a \textit{classification task} by dividing the Earth into discrete geographic grids. These models are trained to predict the specific grid cell containing the query image, outputting its center coordinate as the final prediction. However, this strategy suffers from inherent errors, as the actual location can be far from the grid center even when the classification is correct. More recent methodologies generally diverge into two main paradigms: retrieval-based methods and generation-based methods. (1) \textit{Retrieval-based methods}~\cite{workman2015wide,liu2019lending,zhu2021vigor,lin2022joint,zhu2022transgeo,zhang2023cross,ye2025cross} treat geolocalization as a similarity search problem. They maintain a reference database of geotagged images~\cite{haas2024pigeon,jia2025georanker,tian2017cross,shi2020looking,zhu2023r2former} or GPS entries~\cite{vivanco2023geoclip} and identify the most similar candidates for a given query image. These approaches often achieve high precision for previously seen landmarks, but they may struggle with unseen views or visually similar yet geographically distant regions. (2) \textit{Generation-based methods}~\cite{fang2026geomr,zhou2024img2loc,jia2024g3,ghasemi2025geotoken} rely on LVLMs to directly infer geographic locations. This paradigm has recently evolved into agentic frameworks, where models utilize internal semantic reasoning~\cite{li2024georeasoner,li2025recognition,wang2025gre,jin2026geoagentlearninggeolocatereinforced} or invoke external search tools~\cite{ji2026thinking} to deduce coordinates from visual clues like climate, architecture, and vegetation. Nevertheless, these models can suffer from spatial hallucinations and struggle to achieve fine-grained spatial precision~\cite{jia2026spotagent}. 
GeoRouter diverges from these existing categories by proposing a new framework that dynamically routes each query to the most suitable paradigm, thereby breaking the performance ceiling of single-model systems.

\subsection{Routing and Dynamic Model Selection}

The concept of dynamic model selection is grounded in the algorithm selection problem, which posits that no single algorithm is universally optimal for all problem instances~\cite{kotthoff2016algorithm,kerschke2019automated}. Instead, distinct algorithms exhibit performance complementarity, where their relative effectiveness varies based on the specific characteristics of the input. Recently, this methodology has gained significant traction in the domain of Large Language Models (LLMs). Known as LLM routing, the objective is to dynamically direct user queries to the most suitable component—typically selecting between a capability-rich proprietary model and a cost-effective open-source counterpart—to optimize resource consumption without compromising response quality~\cite{varangot2025doing}. Existing literature primarily categorizes these approaches into four streams: (1) \textit{Similarity-based Routing Methods}~\cite{jang2023exploring,ong2024routellm} route queries by clustering vector embeddings to identify semantically similar historical interactions. (2) \textit{Supervised Routing Methods}~\cite{zhuang2024embedllm,ding2024hybrid,hu2024routerbench} train standalone classifiers on query-performance profiles to explicitly predict the optimal model for specific domains. (3) \textit{Reinforcement Learning-based Routing Methods}~\cite{sikeridis2025pickllm,lu2024routing} formulate routing as a sequential decision-making process, learning optimal policies through trial-and-error feedback. (4) \textit{Generative Routing Methods}~\cite{li2024retrieval,patil2024gorilla} leverage LLM reasoning capabilities to directly generate routing decisions via instruction tuning or in-context learning.
GeoRouter is the first framework to dynamically route between retrieval and generation paradigms, synergizing their complementary strengths to achieve superior localization accuracy.

%% file: 3Methodology.tex
\section{Methodology}

\begin{figure}
    \centering
    \includegraphics[width=\linewidth]{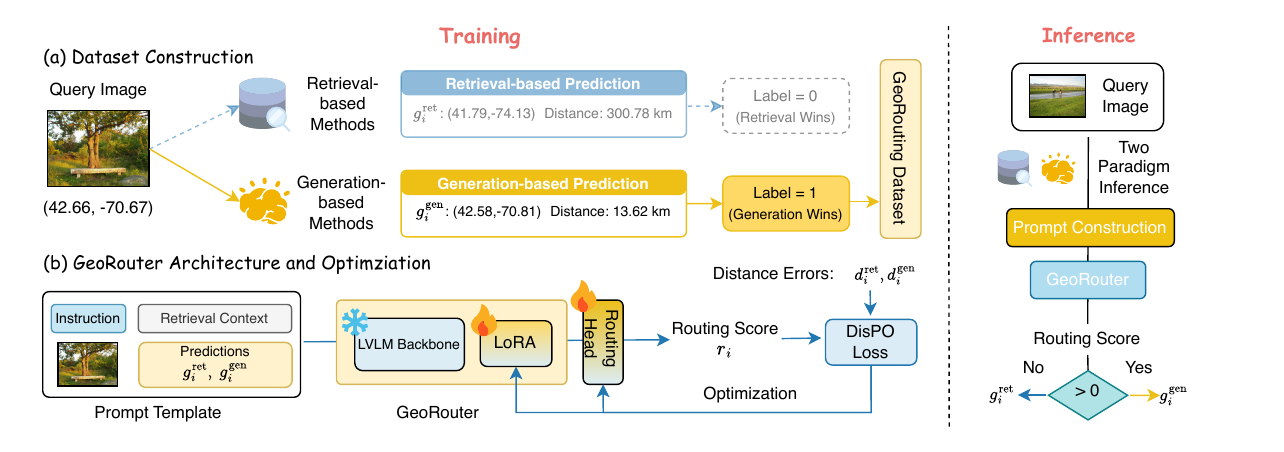}
    \caption{Overview of the dynamic routing framework -- GeoRouter.}
    \label{fig:overview}
\end{figure}

In this section, we present the details of GeoRouter, our dynamic routing framework for worldwide image geolocalization. Figure~\ref{fig:overview} provides an overview of the framework, which consists of two stages: training and inference. The training stage begins with dataset construction in Section~\ref{sec:dataset}, where we build the GeoRouting dataset to support routing supervision. It then proceeds to the model architecture in Section~\ref{sec:architecture} and the optimization in Section~\ref{sec:optimization}.
During the inference stage (Section~\ref{sec:inference}), GeoRouter analyzes the input image and predicts a routing score to select the most suitable paradigm for each query.

\subsection{Dataset Construction} \label{sec:dataset}

Training GeoRouter requires supervision that reflects the relative performance of the two paradigms on each query. We construct this supervision based on existing geotagged image databases (e.g., MP16~\cite{DBLP:conf/mediaeval/ChoiHLT16}), from which we sample a set of query images. Each sampled query consists of an image $I_i$ and its ground-truth geographic coordinate $g_i$. For each query $I_i$, we generate spatial predictions using two distinct paradigms: a retrieval-based method $\mathcal{R}$ and a generation-based method $\mathcal{G}$. The retrieval method extracts a set of similar candidates $\mathcal{C}_i = \{c_{i}^1, \dots, c_{i}^K\} = \mathcal{R}(I_i)$, where each candidate $c_i^k$ denotes a database entry associated with an image $I_{c_i^k}$ and its geographic coordinate $\text{gps}_{c_i^k}$. The location of the top-ranked candidate $c_i^1$ is taken as the retrieval prediction:$g_i^{\text{ret}} = \text{gps}_{c_i^1}$.
Concurrently, the generation method employs an LVLM to directly infer a coordinate prediction: $g_i^{\text{gen}}=\mathcal{G}(I_i)$. We calculate the geolocalization error for both methods against the ground-truth coordinate as $d_i^{\text{ret}} = \mathcal{D}(g_i^{\text{ret}}, g_i)$ and $d_i^{\text{gen}} = \mathcal{D}(g_i^{\text{gen}}, g_i)$, where $\mathcal{D}(\cdot,\cdot)$ represents the geographic distance function.
The binary routing label is defined as $y_i = \mathbf{1}[d_i^{\text{gen}} < d_i^{\text{ret}}]$, 
where $y_i = 1$ indicates generation is preferred and $y_i = 0$ indicates retrieval is preferred.
Consequently, we format each training instance in the GeoRouting dataset as the tuple $(I_i, g_i^{\text{ret}}, g_i^{\text{gen}}, d_i^{\text{ret}}, d_i^{\text{gen}}, \mathcal{C}_i, y_i)$ to provide comprehensive supervision for the routing model.

\subsection{GeoRouter} \label{sec:architecture}

Figure~\ref{fig:overview}(b) illustrates the overall architecture of GeoRouter. Existing methods typically follow a single paradigm, either relying entirely on retrieval-based matching~\cite{haas2024pigeon,vivanco2023geoclip,jia2025georanker} or solely on generation-based reasoning~\cite{li2025recognition}, which overlooks their complementary strengths and limits overall performance. To address this limitation, GeoRouter introduces a dynamic routing mechanism that formulates paradigm selection as a learnable decision problem. Instead of directly predicting geographic coordinates, GeoRouter evaluates the query image and estimates which paradigm is more reliable for the given input.

\textbf{Prompt Template.} To enable the LVLM to perform paradigm selection, the query image and auxiliary contextual information are assembled into a unified prompt template. The prompt explicitly frames routing as a decision task and presents predictions from two paradigms and candidate information as reference inputs. The prompt template is shown as follows:

\begin{tcolorbox}[colback=gray!5!white,colframe=gray!75!black]
Task: Decide whether to use generation or retrieval for this image's geolocalization. Query: 
\textcolor{myblue}{\{Query Image $I_i$\}}  Generation-based Prediction: \textcolor{myblue}{\{Generated GPS Coordinate $g_i^{\text{gen}}$\}}  Retrieval-based Prediction: \textcolor{myblue}{\{Top-1 Retrieved GPS Coordinate $g_i^{\text{ret}}$\}}\textcolor{myblue}{\{Top-1 Retrieved Image $I_{c_i^1}$\}}  Other Retrieved Candidate Coordinates:  
\textcolor{myblue}{\{Top-K Retrieved GPS List (excluding top-1) $\{\text{gps}_{c_i^k}\}_{k=2}^{K}$\}}
\end{tcolorbox}

Formally, the assembled prompt is denoted as $\mathbf{x}_i = \mathcal{P}(I_i, g_i^{\text{gen}}, g_i^{\text{ret}}, \mathcal{C}_i)$, where $\mathcal{P}(\cdot)$ represents the prompt construction function. This structured design allows the model to jointly consider visual content, predicted coordinates, and retrieval context when estimating the relative reliability of the two paradigms. It is worth noting that variants of this template under different input settings (e.g., using only the query image) are also evaluated, with a detailed analysis provided in Section~\ref{sec:ablation}.

\textbf{Architecture.} The constructed prompt input $\mathbf{x}_i$ is fed into a pretrained LVLM backbone to jointly encode visual and textual information. To adapt the model to the routing task in a parameter-efficient manner, Low-Rank Adaptation (LoRA) modules~\cite{hu2022lora} are injected into selected transformer layers during training, while the majority of backbone parameters remain frozen.

We use the hidden state of the last token in the final layer as the joint representation of the input: 
Let $\mathbf{u}_i = \text{LVLM}(\mathbf{x}_i)_{[-1]} \in \mathbb{R}^{m}$, where $m$ is the hidden dimension. A linear routing head then projects $\mathbf{u}_i$ to a scalar value by $r_i = \boldsymbol{\theta}^\top \mathbf{u}_i$, where $\boldsymbol{\theta} \in \mathbb{R}^{m}$ is the trainable parameter of the routing head. The resulting scalar $r_i \in \mathbb{R}$ indicates the model’s preference between the retrieval-based and generation-based paradigms.

\subsection{Optimization} \label{sec:optimization}

To effectively train GeoRouter, relying on standard binary classification with hard labels ignores the continuous nature of geographic errors. Therefore, we introduce a distance-aware optimization method that converts the geographic distance disparity between the two paradigms into continuous soft labels. This strategy allows the model to learn not only which paradigm performs better but also the magnitude of the performance gap, leading to more calibrated and robust routing decisions. 

\textbf{Distance-Aware Preference Optimization.} Given a training instance $\mathbf{x}_i$, let $d_i^{\text{ret}}$ and $d_i^{\text{gen}}$ denote the geographic prediction errors for the retrieval and generation paradigms, respectively. Instead of using the discrete binary label, we compute a continuous preference score. First, we calculate the logarithmic difference between the two errors to capture their relative scale: 
\begin{equation}
    \Delta_i = \log(d_i^{\text{ret}} + \epsilon) - \log(d_i^{\text{gen}} + \epsilon),
\end{equation}
where $\epsilon$ is a small constant included to ensure numerical stability. Next, we apply a sigmoid function to transform this unbounded difference into a soft probability label $p_i \in (0, 1)$:
\begin{equation}
    p_i = \sigma(\alpha \cdot \Delta_i) = \frac{1}{1 + \exp(-\alpha \cdot \Delta_i)},
\end{equation}
where $\alpha$ is a scaling hyperparameter that controls the steepness of the soft label assignment. Under this formulation, when the generation method is significantly more accurate than the retrieval method ($d_i^{\text{gen}} \ll d_i^{\text{ret}}$), $\Delta_i$ becomes a large positive value, pushing $p_i$ toward $1$. Conversely, when the retrieval method is superior, $\Delta_i$ becomes negative, and $p_i$ approaches $0$. According to Section~\ref{sec:architecture}, $r_i$ denotes the scalar logit predicted by the routing head of GeoRouter for the $i$-th query. We optimize the model using a weighted binary cross-entropy loss function computed over the soft labels: $\mathcal{L} = - \frac{1}{N} \sum_{i=1}^{N} \left[ p_i \log(\sigma(r_i)) + (1 - p_i) \log(1 - \sigma(r_i)) \right]$,
where $N$ is the batch size and $\sigma(r_i)$ represents the predicted probability of selecting the generation-based paradigm. By directly minimizing this distance-aware objective, the framework explicitly aligns the model's continuous confidence with the actual geographic performance gap between the two paradigms.

\subsection{Inference} \label{sec:inference}

During the inference stage, the primary objective is to dynamically assign each query to the most suitable paradigm to maximize accuracy. Given a new query image $I_i$, the framework first collects its associated contextual information, including the retrieval-based prediction $g_i^{\text{ret}}$, the generation-based prediction $g_i^{\text{gen}}$, and the retrieved candidate set $\mathcal{C}_i$, to construct the input prompt $\mathbf{x}_i$. GeoRouter then processes this prompt to compute a continuous routing score by $r_i = \text{GeoRouter}(\mathbf{x}_i)$ and assigns the final 
prediction as $\hat{g}_i = g_i^{\text{gen}}$ if $r_i > 0$, 
and $\hat{g}_i = g_i^{\text{ret}}$ otherwise. 
The ultimate geographic prediction for a set of queries is the aggregation of these individually routed coordinates, ensuring that the system leverages the complementary strengths of both paradigms.

%% file: 4Experiments.tex
\section{Experiments}

\subsection{Experimental Setup}

\textbf{Datasets and Evaluation Metrics.} To build the GeoRouting dataset for training, we adopt the MP16-Pro database~\cite{jia2024g3} as our source of reference images and geographic coordinates. For testing, we align our evaluation with standard geolocalization literature~\cite{vivanco2023geoclip,jia2025georanker,zhou2024img2loc} and evaluate GeoRouter on two widely used benchmarks: IM2GPS3K~\cite{hays2008im2gps} and YFCC4K~\cite{thomee2016yfcc100m}. To quantify prediction accuracy, we compute the geodesic distance between the predicted locations and the ground-truth coordinates. The overall performance is reported as the percentage of queries successfully localized within five predefined distance thresholds: 1km, 25km, 200km, 750km, and 2500km. The routing accuracy is also analyzed in Section~\ref{sec:routing}.

\textbf{GeoRouting Dataset.} To support the training and evaluation of routing models in worldwide image geolocalization, we construct the GeoRouting dataset. Specifically, we sample \textbf{100K instances} from the MP16-Pro database to serve as the queries. For each query, we generate predictions using both the retrieval-based and generation-based paradigms. We then compute the distance between each prediction and the ground-truth coordinate to determine a binary routing label indicating the more suitable paradigm. Consequently, each complete data sample contains the query image data, ground-truth GPS, retrieved candidate context, predictions from both paradigms, their respective distance errors, and the binary routing label. By publicly releasing this dataset, we aim to provide a standard benchmark that advances research in geolocalization routing and multimodal decision-making tasks. Detailed examples of the data entries are provided in \textbf{Appendix~\ref{sec:appendix_georouting}} for reference.

\textbf{Implementation Details.} To represent the generation-based paradigm, we select Gemini-2.5-Flash\footnote{\url{https://docs.cloud.google.com/vertex-ai/generative-ai/docs/models/gemini/2-5-flash}} due to its strong performance on existing image geolocalization benchmarks~\cite{li2025pixels,jia2025geoarena}. For the retrieval-based paradigm, we adopt the current state-of-the-art retrieval-based method, GeoRanker~\cite{jia2025georanker}, utilizing its default configurations without generated candidates. We implement the LVLM backbone of GeoRouter using Qwen2-VL-7B-Instruct\footnote{\url{https://huggingface.co/Qwen/Qwen2-VL-7B-Instruct}}. During training, we apply LoRA to efficiently fine-tune the backbone, specifically targeting the \texttt{q\_proj}, \texttt{k\_proj}, and \texttt{v\_proj} modules. We configure LoRA with a rank of 16, a scaling factor of 32, and a dropout rate of 0.05. GeoRouter is optimized using the AdamW optimizer~\cite{loshchilov2017decoupled} with a learning rate of 1e-4 and a batch size of 24, and it is trained for 3 epochs. For the distance-aware preference optimization, we set the numerical stability constant $\epsilon$ to 1e-6 and the scaling hyperparameter $\alpha$ to $1.6$. More details are given in \textbf{Appendix~\ref{sec:appendix_more_settings}}.

\textbf{Baselines.} To evaluate the effectiveness of our approach, we conduct comparative experiments with 14 baseline methods: [L]kNN, sigma=4~\cite{vo2017revisiting}, PlaNet~\cite{seo2018cplanet}, CPlaNet~\cite{seo2018cplanet}, ISNs~\cite{muller2018geolocation}, Translocator~\cite{pramanick2022world}, GeoDecoder~\cite{clark2023we}, GeoCLIP~\cite{vivanco2023geoclip}, Img2Loc~\cite{zhou2024img2loc}, PIGEON~\cite{haas2024pigeon}, G3~\cite{jia2024g3}, GeoToken~\cite{ghasemi2025geotoken}, GeoRanker~\cite{jia2025georanker}, GeoBayes~\cite{shi2026geobayes}, and Geo-R~\cite{wu2026vision}. Detailed descriptions of baselines are provided in \textbf{Appendix~\ref{sec:appendix_baselines}}.

\subsection{Main Results}

\begin{table}[t]
\centering
\caption{\textbf{Main results} on the IM2GPS3K and YFCC4K benchmarks. Higher values indicate better performance. The best and second-best results are highlighted in \textbf{bold} and \underline{underlined}, respectively. $\Delta$ denotes the relative improvement of GeoRouter over the best baseline.}
\label{tab:overall}
\resizebox{\linewidth}{!}{
\begin{tabular}{lccccccccccc} 
\toprule
\multicolumn{2}{c}{\multirow{2}{*}{\textbf{Methods}}} & \multicolumn{5}{c}{IM2GPS3K}                                                                                                                                                                                                                                                          & \multicolumn{5}{c}{YFCC4K}                                                                                                                                                                                                                                                             \\ 
\cmidrule[\heavyrulewidth]{3-12}
\multicolumn{2}{c}{}                         & \begin{tabular}[c]{@{}c@{}}Street\\1km\end{tabular} & \begin{tabular}[c]{@{}c@{}}City\\25km\end{tabular} & \begin{tabular}[c]{@{}c@{}}Region\\200km\end{tabular} & \begin{tabular}[c]{@{}c@{}}Country\\750km\end{tabular} & \begin{tabular}[c]{@{}c@{}}Continent\\2500km\end{tabular} & \begin{tabular}[c]{@{}c@{}}Street\\1km\end{tabular} & \begin{tabular}[c]{@{}c@{}}City\\25km\end{tabular} & \begin{tabular}[c]{@{}c@{}}Region\\200km\end{tabular} & \begin{tabular}[c]{@{}c@{}}Country\\750km\end{tabular} & \begin{tabular}[c]{@{}c@{}}Continent\\2500km\end{tabular}  \\ 
\midrule
\text{[L]kNN}, sigma=4~\cite{vo2017revisiting} & ICCV'17                    & 7.2                                                 & 19.4                                               & 26.9                                                  & 38.9                                                   & 55.9                                                      & 2.3                                                 & 5.7                                                & 11.0                                                    & 23.5                                                   & 42.0                                                         \\
PlaNet~\cite{weyand2016planet}         & ECCV'16                    & 8.5                                                 & 24.8                                               & 34.3                                                  & 48.4                                                   & 64.6                                                      & 5.6                                                 & 14.3                                               & 22.2                                                  & 36.4                                                   & 55.8                                                       \\
CPlaNet~\cite{seo2018cplanet}       & ECCV'18                    & 10.2                                                & 26.5                                               & 34.6                                                 & 48.6                                                   & 64.6                                                     & 7.9                                                 & 14.8                                               & 21.9                                                  & 36.4                                                   & 55.5                                                       \\
ISNs~\cite{muller2018geolocation}           & ECCV'18                    & 10.5                                                & 28.0                                                 & 36.6                                                  & 49.7                                                   & 66.0                                                        & 6.5                                                 & 16.2                                               & 23.8                                                  & 37.4                                                   & 55.0                                                         \\
Translocator~\cite{pramanick2022world}   & ECCV'22                    & 11.8                                                & 31.1                                               & 46.7                                                  & 58.9                                                   & 80.1                                                      & 8.4                                                 & 18.6                                               & 27.0                                                    & 41.1                                                   & 60.4                                                       \\
GeoDecoder~\cite{clark2023we}     & ICCV'23                    & 12.8                                                & 33.5                                               & 45.9                                                  & 61.0                                                     & 76.1                                                      & 10.3                                                & 24.4                                               & 33.9                                                  & 50.0                                                     & 68.7                                                       \\
GeoCLIP~\cite{vivanco2023geoclip}        & NeurIPS'23                 & 14.11                                               & 34.47                                              & 50.65                                                 & 69.67                                                  & 83.82                                                     & 9.59                                                & 19.31                                              & 32.63                                                 & 55.00                                                     & 74.69                                                      \\
Img2Loc~\cite{zhou2024img2loc}        & SIGIR'24                   & 15.34                                               & 39.83                                              & 53.59                                                 & 69.70                                                   & 82.78                                                     & 19.78                                       & 30.71                                      & 41.40                                          & 58.11                                                  & 74.07                                                      \\
PIGEON~\cite{haas2024pigeon}         & CVPR'24                    & 11.30                                                & 36.70                                               & 53.80                                                  & 72.40                                           & 85.30                                              & 10.4                                                & 23.70                                               & 40.60                                                  & 62.20                                           & 77.70                                               \\
G3~\cite{jia2024g3}              & NeurIPS'24                 & 16.65                                       & 40.94                                      & 55.56                                         & 71.24                                                  & 84.68                                                     & 23.99                                      & 35.89                                     & 46.98                                        & 64.26                                         & 78.15                                             \\
GeoToken~\cite{ghasemi2025geotoken}              & ICDM'25                 &   16.80                                      & 39.60                                     & 53.80                                        & 70.80                                         & 85.00                                            &      24.30                                               &    35.30                                                &    46.60                                                   &    64.20                                                    &      78.60                                                      \\
GeoRanker~\cite{jia2025georanker}              & NeurIPS'25                 &   \uline{18.79}                                      & \uline{45.05}                                     & \uline{61.49}                                        & \uline{76.31}                                         & \uline{89.29}                                            &      \uline{32.94}                                               &    \uline{43.54}                                                &    \uline{54.32}                                                   &    \uline{69.79}                                                    &      \uline{82.45}                                                      \\
GeoBayes~\cite{shi2026geobayes} & AAAI'26 & 6.30 & 34.70 & 53.60 & 73.70 & 85.90 & 4.90 & 16.10 & 30.90 & 55.80 & 75.40 \\
Geo-R~\cite{wu2026vision} & AAAI'26 & 18.10 & 41.53 & 58.31 & 75.33 & 86.42 & 10.47 & 22.67 & 40.04 & 60.83 & 75.84 \\
\midrule
\rowcolor{gray!15} \textbf{GeoRouter}            &       Ours                     & \textbf{20.82}                                      & \textbf{50.48}                                     & \textbf{65.73}                                        & \textbf{80.35}                                         & \textbf{90.66}                                            &      \textbf{32.98}                                               &    \textbf{46.01}                                                &    \textbf{57.52}                                                   &    \textbf{72.02}                                                    &      \textbf{83.02}                                                      \\ \rowcolor{gray!15}
Rel. Improvement        &      $\Delta$                      & $\uparrow10.80\%$                                     & $\uparrow12.05\%$                                    & $\uparrow6.90\%$                                       & $\uparrow5.29\%$                                        & $\uparrow1.53\%$                                           &      $\uparrow0.12\%$                                               &      $\uparrow5.67\%$                                              &   $\uparrow5.89\%$                                                    &  $\uparrow3.20\%$                                                      &   $\uparrow0.69\%$                                                         \\
\bottomrule
\end{tabular}}
\end{table}

To validate the effectiveness of the proposed routing framework, we compare GeoRouter against state-of-the-art geolocalization methods on the IM2GPS3K and YFCC4K benchmarks. As shown in Table~\ref{tab:overall}, the experimental results reveal the following main observations. GeoRouter consistently achieves the highest accuracy across all distance thresholds on both datasets. Specifically, GeoRouter demonstrates its most notable advantages at the city-level spatial scale, outperforming the best baselines by relative margins of 12.05\% and 5.67\% at the 25km threshold on the IM2GPS3K and YFCC4K datasets, respectively. The results confirm the limitations of static models. Relying exclusively on either retrieval matching (e.g., GeoCLIP, GeoRanker) or semantic generation (e.g., Img2Loc, G3) restricts the overall spatial accuracy. By explicitly evaluating the confidence gap and assigning the query to the optimal paradigm, GeoRouter effectively overcomes this performance ceiling. In summary, GeoRouter improves worldwide image geolocalization accuracy by combining the complementary strengths of the retrieval and generation paradigms. Additionally, \textbf{Appendix~\ref{sec:appendix_qualitative_examples}} provides qualitative examples categorized by distance thresholds to illustrate the distribution of queries at varying levels of geolocalization accuracy.

\subsection{Ablation Study} \label{sec:ablation}

To systematically evaluate the contribution of each component in GeoRouter, we conduct an ablation study on both datasets. We define the following variants of our approach: (1) \textbf{w/o} DisPO: Our method optimized using a standard Binary Cross-Entropy (BCE) loss with discrete hard labels. (2) \textbf{w/o} $\mathcal{C}$: Our method where the auxiliary retrieved candidate information is removed from the input prompt. (3) \textbf{w/o} $\mathcal{C}$ \& $g^{\text{ret}}$: Our method where both the candidate set and the retrieval-based prediction are removed. (4) \textbf{w/o} $g^{\text{gen}}$: Our method where the generation-based prediction is excluded from the prompt. (5) \textbf{w/o} Context: Our method where all paradigm predictions and candidate information are removed; the routing decision is based solely on the query image.
\setlength{\intextsep}{5pt}
\setlength{\columnsep}{10pt}
\begin{wraptable}{r}{0.5\textwidth}
\centering
\caption{Ablation study on IM2GPS3K.}
\label{tab:ablation}
\resizebox{\linewidth}{!}{
\begin{tabular}{lcccccc} 
\toprule
Methods                   & \begin{tabular}[c]{@{}c@{}}Street\\1km\end{tabular} & \begin{tabular}[c]{@{}c@{}}City\\25km\end{tabular} & \begin{tabular}[c]{@{}c@{}}Region\\200km\end{tabular} & \begin{tabular}[c]{@{}c@{}}Country\\750km\end{tabular} & \begin{tabular}[c]{@{}c@{}}Continent\\2500km\end{tabular} & Average  \\ 
\midrule
\textbf{w/o} DisPO                 & \textbf{20.85}                                               & \uline{49.38}                                              & 64.26                                                 & 79.05                                                  & 89.76                                                     & 60.66    \\
\textbf{w/o} $\mathcal{C}$             & 20.69                                               & 49.18                                              & \uline{64.60}                                                 & \uline{79.75}                                                  & \uline{90.09}                                                     & \uline{60.86}    \\
\textbf{w/o} $\mathcal{C}$ \& $g^{\text{ret}}$             & 19.22                                               & 46.11                                              & 61.76                                                 & 77.11                                                  & 89.29                                                     & 58.69    \\
\textbf{w/o} $g^{\text{gen}}$            & 20.22                                               & 48.31                                              & 64.06                                                 & 78.75                                                  & 89.92                                                     & 60.25    \\
\textbf{w/o} Context & 19.29                                               & 45.85                                              & 61.46                                                 & 77.41                                                  & 89.16                                                     & 58.63    \\
\midrule
Ours                      & \uline{20.82}                                               & \textbf{50.48}                                              & \textbf{65.73}                                                 & \textbf{80.35}                                                  & \textbf{90.66}                                                     & \textbf{61.60}    \\

\bottomrule
\end{tabular}}
\end{wraptable}
The quantitative results on IM2GPS3K are presented in Table~\ref{tab:ablation} and the results for YFCC4K are provided in \textbf{Appendix~\ref{sec:appendix_ablation}}. From the results we can draw several key observations. (1) The full GeoRouter framework achieves the highest average accuracy, confirming that every proposed component contributes positively to the final performance. 
(2) Replacing the DisPO with a standard BCE loss (w/o DisPO) leads to a performance drop, particularly at the intermediate and coarse spatial levels (e.g., 25km and 200km). This demonstrates that discrete binary labels fail to capture the geographic magnitude of prediction errors. By utilizing distance-aware optimization, the model learns a more calibrated confidence score, which is essential for accurate paradigm selection. 
(3) Removing contextual information from the prompt consistently decreases performance. Excluding the candidate set (w/o $\mathcal{C}$) causes a minor drop in accuracy, which indicates that the retrieved candidates provide helpful context for the model to learn routing. Removing either the retrieval contexts (w/o $\mathcal{C}$ \& $g^{\text{ret}}$) or the generation prediction (w/o $g^{\text{gen}}$) leads to larger accuracy reductions. Finally, the variant that relies only on the query image (w/o Context) yields the lowest overall performance. These results demonstrate that providing the predicted coordinates allows the model to compare the paradigms directly, rather than making routing decisions without references.

\subsection{Effect of the Scaling Hyperparameter $\alpha$}

\begin{figure}[h]
    \centering
    \includegraphics[width=\linewidth]{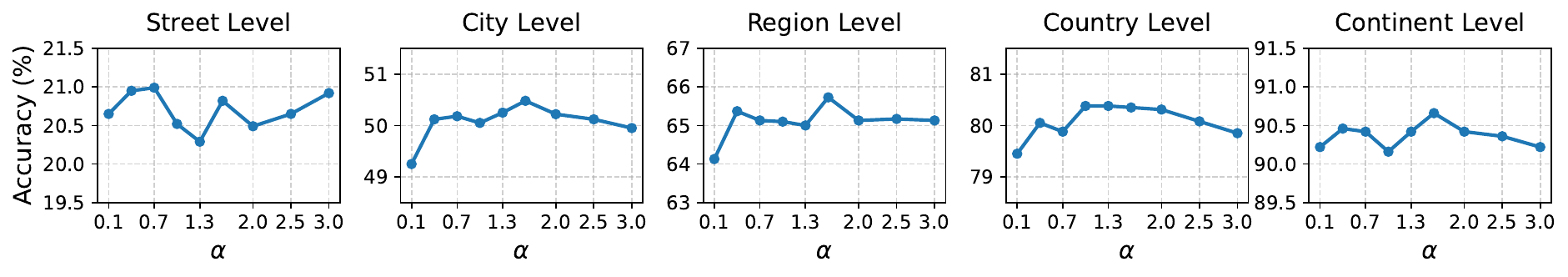}
    \caption{Effect of the hyperparameter $\alpha$ on geolocalization accuracy. The subfigures present the results evaluated on the IM2GPS3K dataset across distance thresholds ranging from 1km to 2500km.}
    \label{fig:hyperparameter}
\end{figure}

To understand the impact of the scaling hyperparameter $\alpha$ within our Distance-Aware objective, we conduct an experiment by systematically varying its value from $0.1$ to $3.0$. The hyperparameter $\alpha$ controls the steepness of the soft probability labels generated from the geographic distance disparity between the retrieval and generation paradigms. We present the geolocalization accuracy across all five distance thresholds on the IM2GPS3K dataset in Figure~\ref{fig:hyperparameter}. The complete evaluation results for both the IM2GPS3K and YFCC4K datasets are provided in \textbf{Appendix~\ref{sec:appendix_hyperparameter}} for reference.

From the visual trends in Figure~\ref{fig:hyperparameter}, we draw several key insights regarding the model behavior. First, setting $\alpha$ to a very low value (e.g., $0.1$) consistently yields the lowest accuracy across most spatial thresholds. A small $\alpha$ produces an overly flat probability distribution, which prevents the model from effectively capturing the magnitude of the performance gap between the two paradigms. Second, as $\alpha$ increases, the geolocalization performance generally improves, reaching an optimal peak around $\alpha = 1.6$. Finally, when $\alpha$ is set too high (e.g., $2.5$ or $3.0$), the accuracy begins to gradually decline. An excessively large $\alpha$ causes the sigmoid transformation to approximate a step function, which effectively reverts the continuous soft labels back into discrete hard binary labels. This degradation highlights the necessity of continuous distance modeling and validates the core design of DisPO.

\subsection{Effectiveness of Dynamic Routing} \label{sec:routing}

\begin{table*}[t]
    \centering
    \caption{Comparison of geolocalization and routing accuracy across different distance thresholds. The best performance among the realistic (non-Oracle) methods is highlighted in \textbf{bold}, while the Oracle results are presented in \textit{italics} as the theoretical upper bound.}
    \label{tab:routing}
    \resizebox{\textwidth}{!}{%
    \begin{tabular}{ll cccccc c cccccc}
        \toprule
        \multirow{2}{*}{Dataset} & \multirow{2}{*}{Method} & \multicolumn{6}{c}{Geolocalization Accuracy (\%)} && \multicolumn{6}{c}{Routing Accuracy (\%)} \\
        \cmidrule{3-8} \cmidrule{10-15}
        & & 1km & 25km & 200km & 750km & 2500km & Average && 1km & 25km & 200km & 750km & 2500km & Average \\
        \midrule
        \multirow{4}{*}{IM2GPS3K} 
        & Pure Retrieval  & 18.42 & 43.14 & 59.29 & 75.31 & 88.36 & 56.90 && 57.28 & 48.44 & 48.48 & 47.57 & 48.37 & 50.03 \\
        & Pure Generation & 17.55 & 47.91 & 63.30 & 78.85 & 88.59 & 59.24 && 42.72 & 51.56 & 51.52 & 52.43 & 51.63 & 49.97 \\
        & GeoRouter       & \textbf{20.82} & \textbf{50.48} & \textbf{65.73} & \textbf{80.35} & \textbf{90.66} & \textbf{61.61} && \textbf{65.52} & \textbf{63.36} & \textbf{62.09} & \textbf{61.61} & \textbf{61.30} & \textbf{62.78} \\
        & \textcolor{gray}{Oracle}          & \textcolor{gray}{\textit{24.29}} & \textcolor{gray}{\textit{54.55}} & \textcolor{gray}{\textit{71.37}} & \textcolor{gray}{\textit{85.09}} & \textcolor{gray}{\textit{93.89}} & \textcolor{gray}{\textit{65.84}} && \textcolor{gray}{\textit{100.00}} & \textcolor{gray}{\textit{100.00}} & \textcolor{gray}{\textit{100.00}} & \textcolor{gray}{\textit{100.00}} & \textcolor{gray}{\textit{100.00}} & \textcolor{gray}{\textit{100.00}} \\
        \midrule
        \multirow{4}{*}{YFCC4K} 
        & Pure Retrieval  & 32.47 & 42.70 & 53.13 & 69.95 & 82.67 & 56.18 && 85.23 & 73.34 & 67.51 & 65.47 & 64.44 & 71.20 \\
        & Pure Generation & 10.43 & 26.85 & 41.93 & 59.30 & 73.28 & 42.36 && 14.77 & 26.66 & 32.49 & 34.53 & 35.56 & 28.80 \\
        & GeoRouter       & \textbf{32.89} & \textbf{46.01} & \textbf{57.52} & \textbf{72.02} & \textbf{83.02} & \textbf{58.29} && \textbf{86.90} & \textbf{80.96} & \textbf{77.44} & \textbf{72.83} & \textbf{70.92} & \textbf{77.81} \\
        & \textcolor{gray}{Oracle}          & \textcolor{gray}{\textit{35.67}} & \textcolor{gray}{\textit{49.78}} & \textcolor{gray}{\textit{62.43}} & \textcolor{gray}{\textit{77.89}} & \textcolor{gray}{\textit{89.09}} & \textcolor{gray}{\textit{62.97}} && \textcolor{gray}{\textit{100.00}} & \textcolor{gray}{\textit{100.00}} & \textcolor{gray}{\textit{100.00}} & \textcolor{gray}{\textit{100.00}} & \textcolor{gray}{\textit{100.00}} & \textcolor{gray}{\textit{100.00}} \\
        \bottomrule
    \end{tabular}%
    }
\end{table*}

To evaluate the effectiveness of the proposed routing framework, we analyze the routing accuracy and the resulting geolocalization performance against static single-paradigm policies (Pure Retrieval and Pure Generation) and a theoretical upper bound (Oracle). As shown in Table~\ref{tab:routing}, we can make three primary observations. First, GeoRouter consistently outperforms both pure baselines across all metrics, proving its effectiveness in identifying the optimal paradigm per query. Second, comparing the two datasets highlights the robustness of GeoRouter across different error distributions. On the IM2GPS3K dataset, the performance of the pure retrieval and pure generation methods is relatively balanced. In contrast, on the YFCC4K dataset, the pure retrieval method exhibits an advantage over the generation method. Even under this imbalanced scenario, GeoRouter successfully adapts to the underlying data distribution and maintains its performance superiority, demonstrating strong robustness. Third, while GeoRouter successfully exceeds the performance limits of single models, a clear gap remains compared to the Oracle bound (e.g., 61.61\% versus 65.84\% average accuracy on IM2GPS3K). This confirms the routing task's potential and highlights policy optimization as a valuable future direction. We also investigated the transferability of GeoRouter in \textbf{Appendix~\ref{sec:appendix_transferability}}.

\subsection{Effect of Backbone Model Scale}

\begin{figure}
\centering
\begin{minipage}{0.48\linewidth}
\centering
\includegraphics[width=\linewidth]{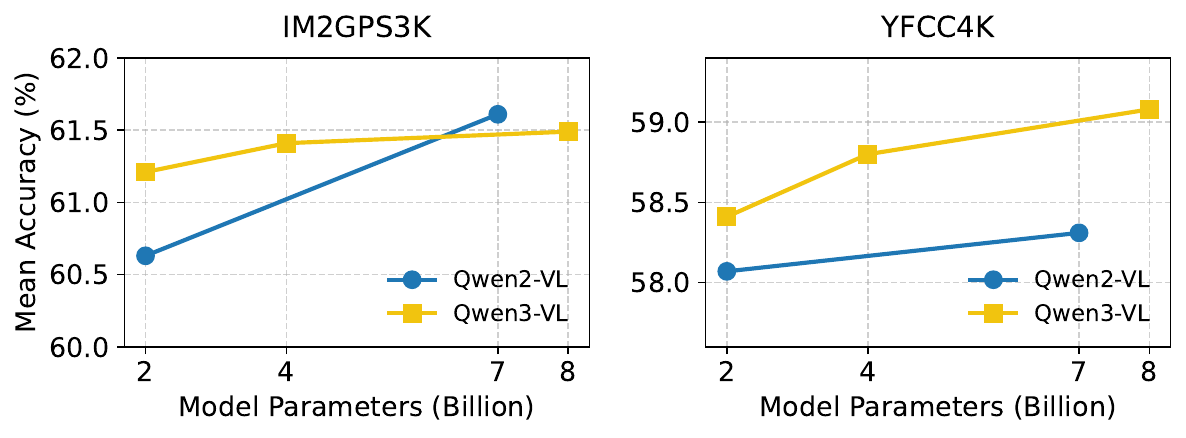}
\captionof{figure}{Effect of backbone model scale on geolocalization performance. The evaluation compares Mean Accuracy across different parameter sizes of the Qwen2-VL and Qwen3-VL families.}
\label{fig:scaling}
\end{minipage}
\hspace{0.01\linewidth}
\begin{minipage}{0.48\linewidth}
\centering
\includegraphics[width=\linewidth]{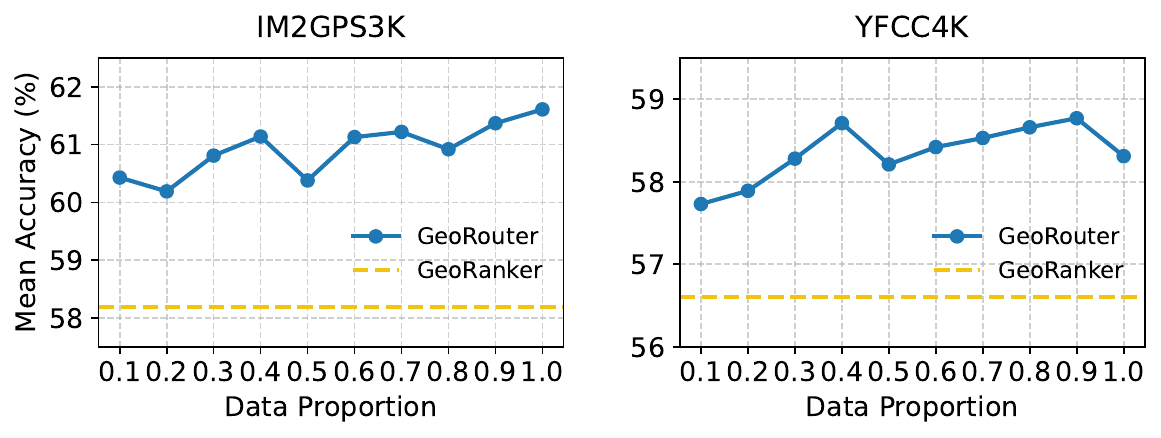}
\captionof{figure}{Data efficiency analysis. The figures illustrate the Mean Accuracy achieved when training GeoRouter with varying proportions of the available dataset.}
\label{fig:data_efficiency}
\end{minipage}
\vspace{-1em}
\end{figure}

To investigate how the parameter scale of the LVLM backbone influences the performance, we evaluate GeoRouter using varying sizes of the Qwen2-VL and Qwen3-VL model families. Figure~\ref{fig:scaling} illustrates the mean accuracy on the IM2GPS3K and YFCC4K datasets across different model scales. From these results, we draw two primary conclusions. First, increasing the number of model parameters consistently improves routing performance. For both the Qwen2-VL and Qwen3-VL families, the mean accuracy exhibits a clear upward trend as the parameter count scales from 2 billion to 7 or 8 billion. 
Second, advancements in the foundational model architecture provide immediate performance benefits. At equivalent or similar parameter scales (e.g., comparing the 2-billion parameter versions), the newer Qwen3-VL models consistently outperform the older Qwen2-VL models on both datasets. This confirms that the GeoRouter framework can easily scale and benefit from ongoing improvements in state-of-the-art LVLM architectures. The details results for each level are provided in \textbf{Appendix~\ref{sec:appendix_scale}}.

\subsection{Data Efficiency Analysis}

To investigate the impact of training data volume on the learned routing policy, we evaluate GeoRouter using varying proportions of the available dataset. As illustrated in Figure~\ref{fig:data_efficiency}, we draw two primary conclusions from the results. First, the mean geolocalization accuracy exhibits a general upward trend as the data proportion increases from 0.1 to 1.0. This indicates that exposure to more training samples steadily improves the routing decisions. Second, the framework demonstrates high data efficiency. Even when restricted to only 10\% of the training data (a proportion of 0.1), GeoRouter consistently surpasses the SOTA baseline GeoRanker on both the IM2GPS3K and YFCC4K datasets. This confirms that the model effectively learns the comparative routing mechanism from a very small fraction of the data, while continuing to benefit from the broader geographic diversity provided by the complete dataset. For detailed experimental results at each distance level and analysis on inference overhead, please refer to the \textbf{Appendix~\ref{sec:appendix_data_efficiency}} and \textbf{Appendix~\ref{sec:appendix_time_efficiency}}.

\subsection{Case Study}

\begin{figure}
    \centering
    \includegraphics[width=\linewidth]{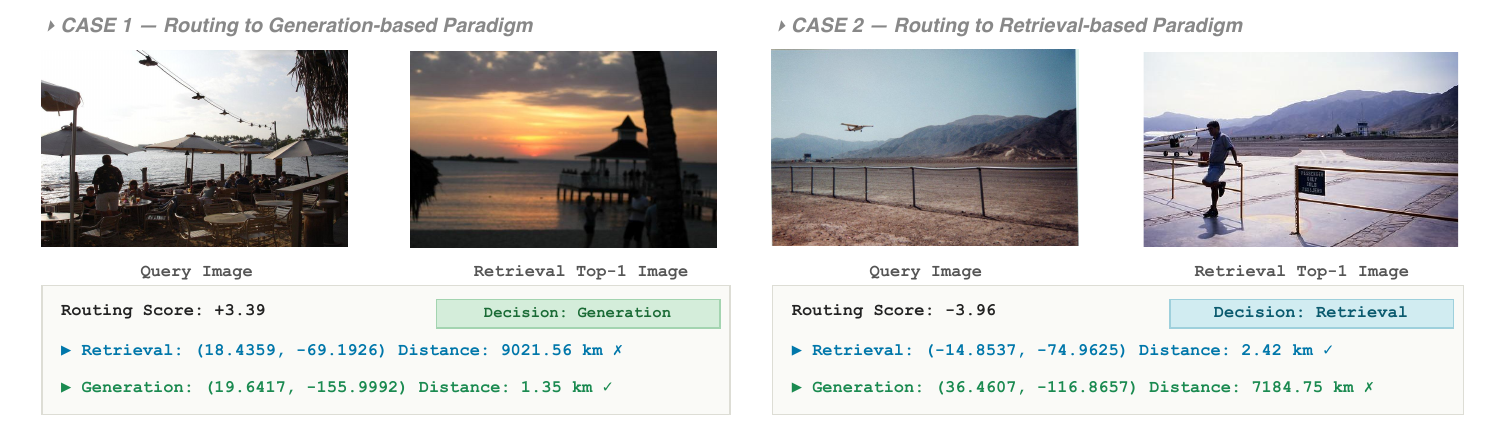}
    \caption{Case studies illustrating GeoRouter's routing decisions across two paradigms.}
    \label{fig:case_study}
\end{figure}

To qualitatively illustrate the routing behavior of GeoRouter, we analyze two representative examples shown in Figure~\ref{fig:case_study}. \textbf{Case 1} presents a coastal dining scene containing generic visual elements. The retrieval method matches a visually similar candidate from a distant location 
(9021.56\,km away), as such scenes offer insufficient discriminative features for precise instance matching. In contrast, the generation-based method correctly infers the geographic context from semantic cues and produces an accurate prediction (1.35\,km). GeoRouter assigns a positive routing score ($r = +3.39$) and correctly selects the generation paradigm. \textbf{Case 2} presents an older photograph of an arid airstrip, where limited image quality and the absence of salient semantic cues make direct coordinate generation 
unreliable (7184.75\,km). However, the distinctive mountain silhouette in the background provides a strong visual fingerprint that enables the retrieval method to identify a geographically accurate candidate (2.42\,km). GeoRouter assigns a negative routing score ($r = -3.96$) and correctly selects the retrieval paradigm.

%% file: 5Conclusion.tex
\section{Conclusion}

In this paper, we propose GeoRouter, a dynamic routing framework that adaptively assigns each query to the most suitable paradigm for worldwide image geolocalization. To support this framework, we introduce Distance-Aware Preference Optimization (DisPO) and construct GeoRouting, the first large-scale dataset designed for geolocalization routing. Extensive experiments on IM2GPS3K and YFCC4K demonstrate that GeoRouter consistently outperforms state-of-the-art baselines.

%% file: 6Appendix.tex
\section{GeoRouting Data Entries} \label{sec:appendix_georouting}

\begin{figure}
    \centering
    \includegraphics[width=\linewidth]{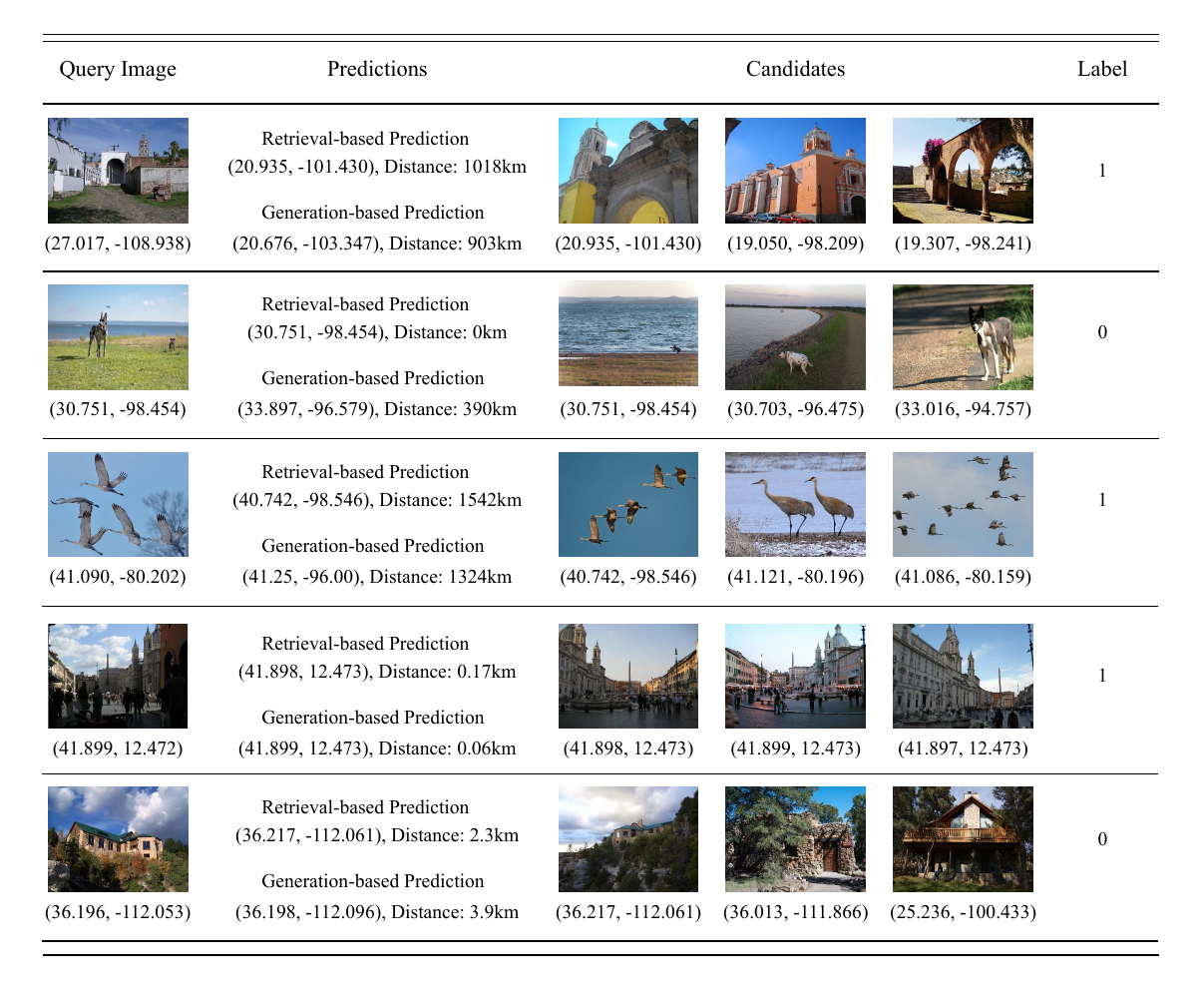}
    \caption{Representative examples of GeoRouting dataset entries. Each row shows a query image with its ground-truth GPS coordinates, predictions from the retrieval-based and generation-based paradigms with their respective distance errors, retrieved candidate images with their GPS coordinates, and the binary routing label (1 = generation-based paradigm is more accurate; 0 = retrieval-based paradigm is more accurate).}
    \label{fig:georouting_examples}
\end{figure}

Each entry in the GeoRouting dataset consists of a query image, its ground-truth GPS coordinates, retrieved candidate images with their coordinates, predictions from both the retrieval-based and generation-based paradigms, the distance error of each prediction, and a binary routing label. The label indicates which paradigm produces the more accurate prediction: label 1 denotes that the generation-based paradigm is more suitable, while label 0 denotes that the retrieval-based paradigm 
is more suitable. For display purposes, Figure~\ref{fig:georouting_examples} shows only three of the ten candidates per entry.

\section{More Details on Experimental Settings} \label{sec:appendix_more_settings}

\begin{table}
\centering
\caption{Detailed experimental configurations.}
\label{tab:appendix_more_settings}
\resizebox{0.62\linewidth}{!}{
\begin{tabular}{lc}
\toprule
Parameter             & Value                                    \\ 
\midrule
Batch Size per Device & 3                                        \\
Batch Size            & 24                                       \\
Dataset Size          & 100K                                     \\
DeepSpeed Stage       & 2                                        \\
GPU                   & NVIDIA H800 $\times$ 8  \\
Training Time         & 50 min / epoch                           \\
Total Parameters      & 8,298,264,064                            \\
Trainable Parameters  & 6,884,864 (0.08\textbackslash{}\%)       \\
Training GPU Memory   & Total 485 GiB, Avg 60.5 GiB / Device     \\
LVLM Backbone         & Qwen2-VL-7B-Instruct                     \\

\bottomrule
\end{tabular}}
\end{table}

Table~\ref{tab:appendix_more_settings} provides a comprehensive summary of the hardware 
and training configuration used in our experiments. All experiments are conducted on a 
node equipped with 8 NVIDIA H800 GPUs. Each training epoch takes approximately 50 minutes, 
and the full 3-epoch training run is completed in under 3 hours, demonstrating the 
computational efficiency of our approach.

\section{Detailed Introduction to Baselines} \label{sec:appendix_baselines}

In this section, we provide introductions to all baseline methods compared in our experiments.

\begin{itemize}[leftmargin=*]
    \item \textbf{$\text{[L]kNN},\sigma=4$}~\cite{vo2017revisiting}. kNN retrieves the top-$k$ most similar images from a reference database and aggregates their GPS coordinates to produce a final prediction. The parameter $\sigma$ controls the spatial bandwidth of the aggregation kernel; as $k$ decreases, the prediction becomes more focused, and at $k=1$ the method reduces to nearest-neighbor retrieval.

    \item \textbf{PlaNet}~\cite{weyand2016planet}. PlaNet is the first method to cast worldwide image geolocalization as a classification problem. It partitions the Earth's surface into geographic cells and trains a convolutional neural network to assign each query image to its corresponding cell. The center coordinate of the predicted cell is used as the final location estimate.

    \item \textbf{CPlaNet}~\cite{seo2018cplanet}. CPlaNet extends PlaNet by introducing a combinatorial partitioning strategy that derives fine-grained output classes through the intersection of multiple coarser geographic partitions. This hierarchical decomposition allows the model to generate more precise location predictions compared to using a single flat partitioning scheme.

    \item \textbf{ISNs}~\cite{muller2018geolocation}. ISNs augment the standard image input with auxiliary scene features that capture high-level environmental context, such as whether a scene is indoor, natural, or urban. Incorporating these additional signals alongside the original image content enables the model to produce richer representations, leading to improved geolocalization accuracy across diverse environments.

    \item \textbf{Translocator}~\cite{pramanick2022world}. Translocator adopts a dual-branch transformer architecture that processes both the original query image and its corresponding semantic segmentation map in parallel. By jointly exploiting appearance and structural information, the model learns more discriminative location-aware features and achieves stronger performance.

    \item \textbf{GeoDecoder}~\cite{clark2023we}. GeoDecoder identifies that prior classification-based methods do not fully exploit the hierarchical nature of geographic information. It addresses this limitation by introducing a cross-attention mechanism that models relationships across features at different spatial granularities, improving the model's capacity to reason about location from complex scene content.

    \item \textbf{GeoCLIP}~\cite{vivanco2023geoclip}. GeoCLIP adapts the CLIP contrastive learning framework for geolocalization by introducing a dedicated GPS encoder that projects geographic coordinates into a shared embedding space alongside image features. This alignment allows the model to directly associate visual content with geographic locations, enabling effective worldwide geolocalization without relying on a fixed classification grid.

    \item \textbf{Img2Loc}~\cite{zhou2024img2loc}. Img2Loc integrates a retrieval-augmented generation (RAG) pipeline into geolocalization. Given a query image, the system first retrieves a set of visually similar reference images and collects their GPS coordinates. These coordinates are then incorporated into a prompt for a large vision-language model, which generates the final location prediction by reasoning over both the visual content and the retrieved geographic context.

    \item \textbf{PIGEON}~\cite{haas2024pigeon}. PIGEON introduces a framework that combines semantically guided geographic cell partitioning, multi-task contrastive pretraining, and a specialized loss function. By grouping candidate locations according to semantic similarity and subsequently applying targeted retrieval to refine predictions, PIGEON achieves notably improved geolocalization accuracy over prior classification and retrieval methods.

    \item \textbf{G3}~\cite{jia2024g3}. G3 is a three-stage RAG-based framework composed of Geo-alignment, Geo-diversification, and Geo-verification. The Geo-alignment stage jointly trains image, GPS, and text encoders to learn location-aware multimodal representations for retrieval. The subsequent stages leverage an LVLM to generate and verify a diverse set of candidate GPS coordinates, with the final prediction selected by comparing candidate similarity against the query using the learned representations.

    \item \textbf{GeoToken}~\cite{ghasemi2025geotoken}. GeoToken reformulates geolocalization as a coarse-to-fine autoregressive token prediction task, inspired by the way humans progressively narrow down a location from a broad region to a specific address. The model uses Google S2 cells as a hierarchical spatial grid and autoregressively predicts location tokens at increasing levels of geographic resolution, conditioned on both visual input and previously predicted tokens.

    \item \textbf{GeoRanker}~\cite{jia2025georanker}. GeoRanker addresses a key limitation of existing two-stage retrieval pipelines, where candidate selection relies on simplistic point-wise similarity heuristics that fail to capture spatial relationships among candidates. GeoRanker employs an LVLM to jointly encode query–candidate pairs and predict geographic proximity. A multi-order distance loss is introduced to supervise both absolute and relative distance rankings, enabling the model to reason about the structured spatial arrangement of candidates and select the most geographically accurate prediction.

    \item \textbf{GeoBayes}~\cite{shi2026geobayes}. GeoBayes is a training-free framework that formulates geo-localization as a Maximum a Posteriori (MAP) estimation problem. A state memory mechanism propagates hypotheses, evidence, and inference context across geographic hierarchy levels, enabling coarse-to-fine geolocalization.

    \item \textbf{Geo-R}~\cite{wu2026vision}. Geo-R is a retrieval-free framework that introduces the Chain of Region paradigm, which decomposes geolocalization into a hierarchical reasoning process. The model is further optimized via reinforcement learning using GRPO with a composite reward combining Haversine distance-based spatial accuracy and output format consistency, enabling spatially grounded refinement of coordinate predictions.

\end{itemize}

\section{Qualitative Examples Across Distance Thresholds} \label{sec:appendix_qualitative_examples}

\begin{figure}
    \centering
    \includegraphics[width=\linewidth]{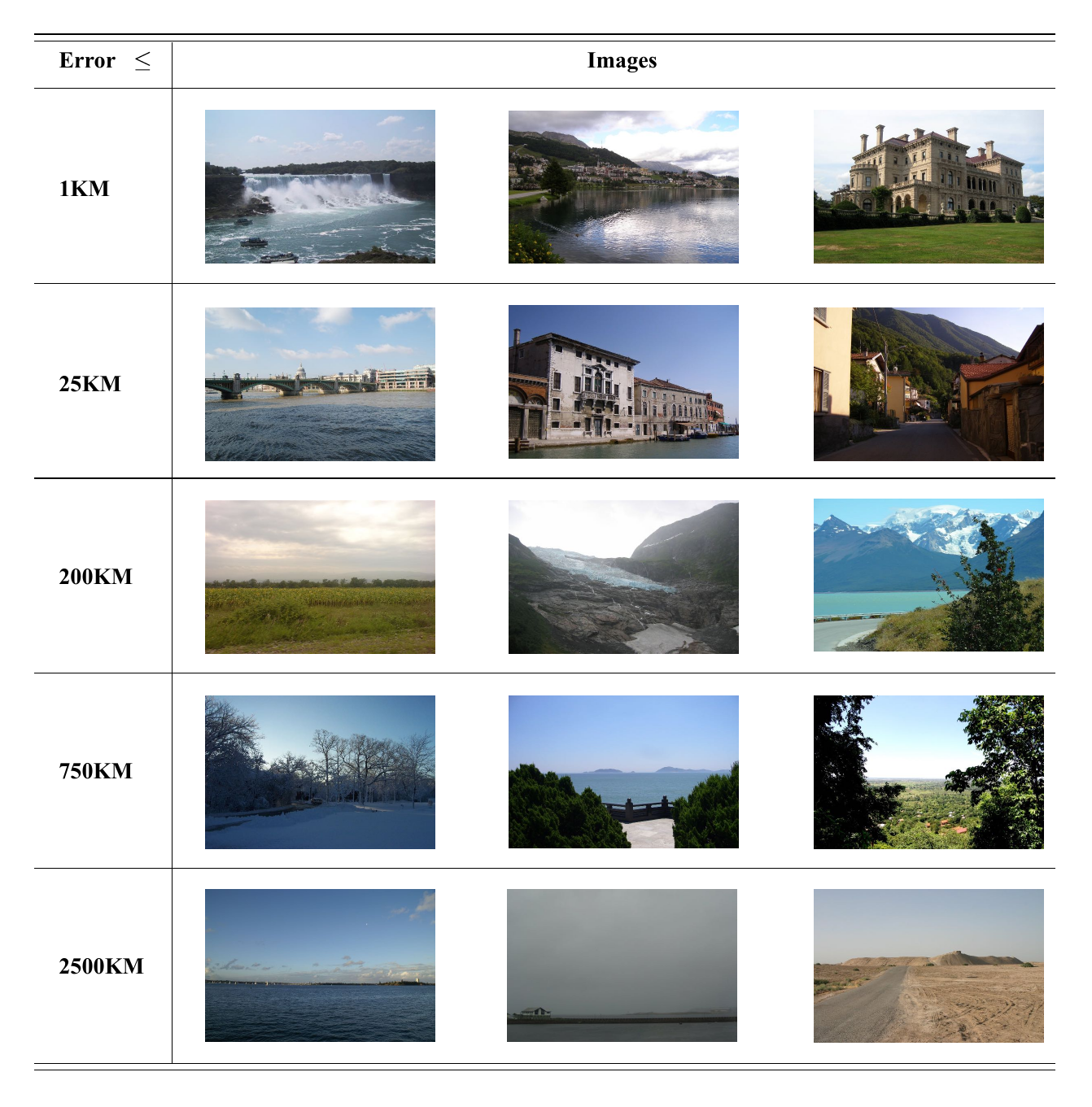}
    \caption{Qualitative examples of GeoRouter predictions across different geolocalization 
accuracy levels}
    \label{fig:appendix_qualitative_examples}
\end{figure}

Figure~\ref{fig:appendix_qualitative_examples} shows representative query images grouped by their geolocalization error across five distance thresholds. Images that are localized within 1km typically contain strong and unambiguous visual identifiers, such as well-known architectural structures or iconic landmarks. These scenes benefit both paradigms: the retrieval method can readily find visually similar database entries for such frequently photographed locations, while the generation method can leverage the world knowledge embedded in LVLMs to reason about them directly. As the error threshold increases to 25km and 200km, the query images tend to feature more generic scenes, such as local churches, rural buildings, or regional landscapes, which provide weaker discriminative cues. At these intermediate levels, accurate geolocalization depends more heavily on subtle contextual signals such as architectural style, vegetation type, or topographic features, making paradigm selection by GeoRouter particularly valuable. At the coarsest thresholds (750km and 2500km), the query images frequently depict scenes with little to no location-specific information, such as open fields, featureless seascapes, or generic natural environments. In these cases, neither the retrieval nor the generation paradigm can reliably infer the true location, and the large errors reflect the fundamental ambiguity inherent in such visually uninformative inputs.

\section{Complete Experimental Results on Ablation Study} \label{sec:appendix_ablation}

\begin{table}
\centering
\caption{Ablation study on IM2GPS3K and YFCC4K.}
\label{tab:appendix_ablation}
\resizebox{\linewidth}{!}{
\begin{tabular}{lcccccccccccc} 
\toprule
\multirow{2}{*}{Methods}                                     & \multicolumn{6}{c}{IM2GPS3K}                                                                                                                                                                                                                                                                           & \multicolumn{6}{c}{YFCC4K}                                                                                                                                                                                                                                                                              \\ 
\cmidrule{2-13}
                                                             & \begin{tabular}[c]{@{}c@{}}Street\\1km\end{tabular} & \begin{tabular}[c]{@{}c@{}}City\\25km\end{tabular} & \begin{tabular}[c]{@{}c@{}}Region\\200km\end{tabular} & \begin{tabular}[c]{@{}c@{}}Country\\750km\end{tabular} & \begin{tabular}[c]{@{}c@{}}Continent\\2500km\end{tabular} & Average        & \begin{tabular}[c]{@{}c@{}}Street\\1km\end{tabular} & \begin{tabular}[c]{@{}c@{}}City\\25km\end{tabular} & \begin{tabular}[c]{@{}c@{}}Region\\200km\end{tabular} & \begin{tabular}[c]{@{}c@{}}Country\\750km\end{tabular} & \begin{tabular}[c]{@{}c@{}}Continent\\2500km\end{tabular} & Average         \\ 
\midrule
\textbf{w/o} DisPO                                           & \textbf{20.85}                                      & \uline{49.38}                                      & 64.26                                                 & 79.05                                                  & 89.76                                                     & 60.66          & \uline{33.05}                                       & \uline{45.75}                                      & \uline{57.08}                                         & 71.65                                                  & 82.80                                                     & 58.06           \\
\textbf{w/o} $\mathcal{C}$                                   & 20.69                                               & 49.18                                              & \uline{64.60}                                         & \uline{79.75}                                          & \uline{90.09}                                             & \uline{60.86}  & \textbf{33.58}                                      & 45.55                                              & 56.42                                                 & \uline{71.94}                                          & \textbf{83.53}                                            & \uline{58.20}   \\
\textbf{w/o} $\mathcal{C}$ \textbackslash{}\& $g^\text{ret}$ & 19.22                                               & 46.11                                              & 61.76                                                 & 77.11                                                  & 89.29                                                     & 58.69          & 29.83                                               & 41.34                                              & 52.05                                                 & 68.67                                                  & 81.50                                                     & 54.67           \\
\textbf{w/o} $g^\text{gen}$                                  & 20.22                                               & 48.31                                              & 64.06                                                 & 78.75                                                  & 89.92                                                     & 60.25          & 33.20                                               & 45.08                                              & 55.62                                                 & 71.25                                                  & 82.94                                                     & 57.61           \\
\textbf{w/o} Context                                         & 19.29                                               & 45.85                                              & 61.46                                                 & 77.41                                                  & 89.16                                                     & 58.63          & 30.14                                               & 41.60                                              & 52.98                                                 & 69.91                                                  & 82.34                                                     & 55.39           \\ 
\midrule
Ours                                                         & \uline{20.82}                                       & \textbf{50.48}                                     & \textbf{65.73}                                        & \textbf{80.35}                                         & \textbf{90.66}                                            & \textbf{61.60} & 32.98                                               & \textbf{46.01}                                     & \textbf{57.52}                                        & \textbf{72.02}                                         & \uline{83.02}                                             & \textbf{58.31}  \\
\bottomrule
\end{tabular}}
\end{table}

Table~\ref{tab:appendix_ablation} presents the complete ablation results across both 
IM2GPS3K and YFCC4K datasets. The trends observed on YFCC4K are consistent with 
those reported on IM2GPS3K in the main paper, confirming that each proposed component 
delivers a positive contribution regardless of the dataset. Regarding the composition of the input prompt, the results highlight a clear hierarchy among 
the contextual components. The two paradigm predictions, $g^{\text{gen}}$ and $g^{\text{ret}}$, 
are the most critical inputs. This indicates that explicitly providing the predicted 
coordinates from both paradigms enables direct comparison and is fundamental to accurate routing 
decisions. By contrast, removing only the retrieved candidate set (\textbf{w/o} $\mathcal{C}$) results 
in a comparatively minor performance degradation, suggesting that while the candidate context 
offers supplementary spatial cues that help the model assess retrieval reliability, the paradigm 
predictions themselves remain the primary basis for routing.

\section{Complete Experimental Results on Hyperparameter Analaysis} \label{sec:appendix_hyperparameter}

\begin{figure}
    \centering
    \includegraphics[width=\linewidth]{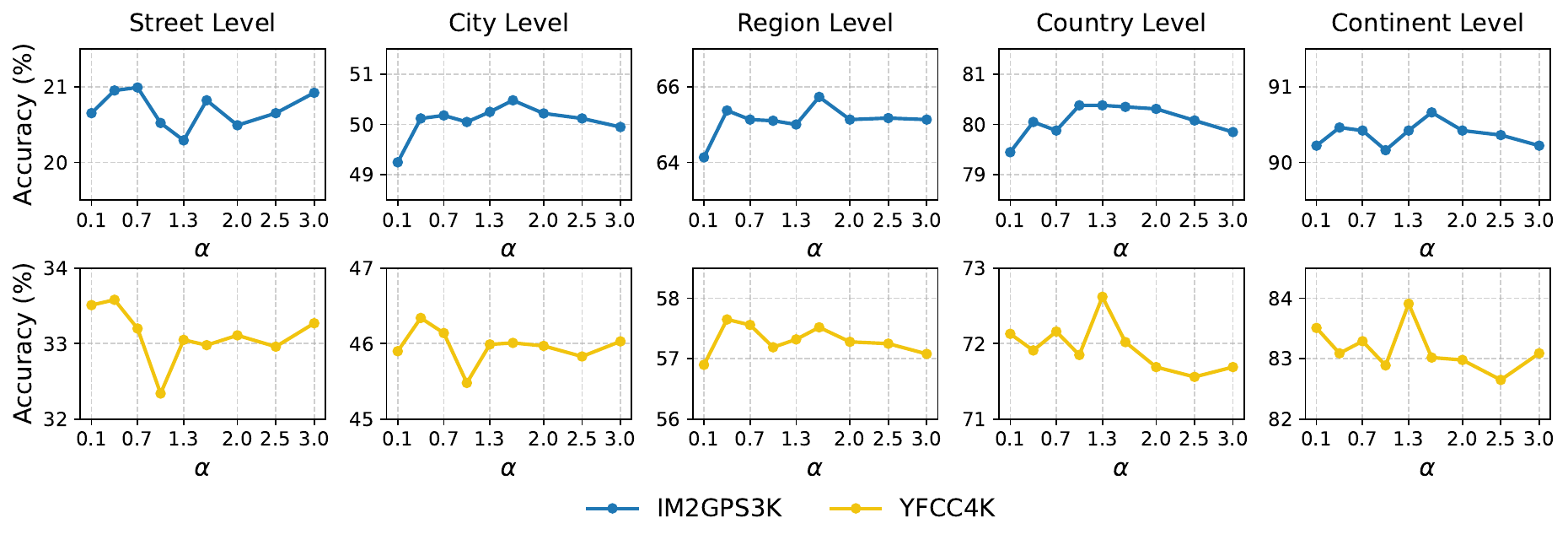}
    \caption{Effect of the hyperparameter $\alpha$ on geolocalization accuracy across all five 
distance thresholds on both IM2GPS3K (blue) and YFCC4K (yellow). Each column corresponds 
to a distance threshold from Street level (1km) to Continent level (2500km).}
    \label{fig:appendix_hyperparameter}
\end{figure}

Figure~\ref{fig:appendix_hyperparameter} presents the complete hyperparameter sensitivity 
results for $\alpha$ on both IM2GPS3K and YFCC4K. On both datasets, setting $\alpha$ to an excessively large value (e.g., $2.5$ or $3.0$) 
leads to a consistent decline in accuracy across most distance thresholds. This behavior is 
particularly evident at the city and region levels (25km and 200km), where calibrated soft 
labels are most critical for distinguishing between the two paradigms. As discussed in the 
main text, an overly large $\alpha$ collapses the continuous soft labels into near-binary 
supervision, undermining the distance-aware modeling that DisPO is designed to provide.

\section{Transferability Analysis} \label{sec:appendix_transferability}

\begin{table}
\centering
\caption{Transferability of the GeoRouter across different generative LVLMs on the IM2GPS3K dataset. All metrics are reported as geolocalization accuracy (\%). The $\Delta$ Mean column indicates the performance gain achieved by integrating our routing mechanism.}
\label{tab:transferability}
\resizebox{0.9\textwidth}{!}{
\begin{tabular}{llccccccc} 
\toprule
\multirow{2}{*}{Base Generative Model}       & \multirow{2}{*}{Method} & \multicolumn{6}{c}{Distance Thresholds}                                                             & \multirow{2}{*}{$\Delta$ Mean}  \\ 
\cmidrule{3-8}
                                             &                         & 1km            & 25km           & 200km          & 750km          & 2500km         & Mean           &                                 \\ 
\midrule
\multirow{2}{*}{Gemini 2.5 Flash}            & Base Model              & 17.55          & 47.91          & 63.30          & 78.85          & 88.59          & 59.24          & \multirow{2}{*}{+4.00\%}          \\
                                             & + GeoRouter             & \textbf{20.82} & \textbf{50.48} & \textbf{65.73} & \textbf{80.35} & \textbf{90.66} & \textbf{61.61} &                                 \\ 
\midrule
\multirow{2}{*}{Gemini 2.5 Flash (Thinking)} & Base Model              & 18.59          & 48.88          & 62.96          & 78.61          & 88.99          & 59.61          & \multirow{2}{*}{+3.37\%}          \\
                                             & + GeoRouter             & \textbf{21.15} & \textbf{50.98} & \textbf{64.83} & \textbf{80.35} & \textbf{90.79} & \textbf{61.62} &                                 \\ 
\midrule
\multirow{2}{*}{Gemini 2.0 Flash}            & Base Model              & 15.42          & 45.38          & 61.23          & 78.14          & 87.92          & 57.62          & \multirow{2}{*}{+4.84\%}          \\
                                             & + GeoRouter             & \textbf{19.45} & \textbf{48.45} & \textbf{63.73} & \textbf{79.88} & \textbf{90.56} & \textbf{60.41} &                                 \\ 
\midrule
\multirow{2}{*}{GPT-5 Mini}                  & Base Model              & 17.95          & 44.84          & 58.66          & 74.21          & 86.49          & 56.43          & \multirow{2}{*}{+4.64\%}          \\
                                             & + GeoRouter             & \textbf{20.95} & \textbf{47.78} & \textbf{61.53} & \textbf{76.58} & \textbf{88.39} & \textbf{59.05} &                                 \\ 
\bottomrule
\end{tabular}}
\end{table}

To evaluate whether the learned routing policy generalizes to different generative models, we conduct a transferability analysis. Specifically, we train GeoRouter using the predictions generated by Gemini 2.5 Flash. During inference, we directly apply this frozen routing model to evaluate predictions from various unseen LVLMs without any additional fine-tuning. The results on the IM2GPS3K dataset, presented in Table~\ref{tab:transferability}, yield several key conclusions. First, GeoRouter consistently improves the geolocalization accuracy of every evaluated LVLM across all distance thresholds. This demonstrates that even when the generation-based method is replaced during inference, the routing mechanism stably enhances the overall system performance. 
Second, the successful transfer across different model families and parameter scales indicates that GeoRouter learns a generalized routing capability rather than overfitting to the specific error distribution of the training LVLM.

\section{Complete Experimental Results on Backbone Model Scale} \label{sec:appendix_scale}

\begin{table}
\centering
\caption{Performance scaling across different sizes of the routing backbone model. The evaluation is conducted using Qwen2-VL and Qwen3-VL families on the IM2GPS3K and YFCC4K datasets. All results are presented as geolocalization accuracy (\%).}
\label{tab:appendix_scaling_law}
\resizebox{\textwidth}{!}{
\begin{tabular}{lcccccccccccc} 
\toprule
\multirow{2}{*}{Backbone Model} & \multicolumn{6}{c}{IM2GPS3K}                   & \multicolumn{6}{c}{YFCC4K}                      \\ 
\cmidrule{2-13}
                                & 1km   & 25km  & 200km & 750km & 2500km & Mean  & 1km   & 25km  & 200km & 750km & 2500km & Mean   \\ 
\midrule
Qwen2-VL-2B                     & 20.35 & 49.02 & 64.36 & 79.35 & 90.06  & 60.63 & 33.49 & 45.50 & 56.97 & 71.38 & 83.00  & 58.07  \\
Qwen2-VL-7B                     & 20.82 & 50.48 & 65.73 & 80.35 & 90.66  & 61.61 & 32.98 & 46.01 & 57.52 & 72.02 & 83.02  & 58.31  \\ 
\midrule
Qwen3-VL-2B                     & 20.95 & 49.68 & 64.80 & 79.91 & 90.72  & 61.21 & 33.55 & 45.90 & 57.30 & 71.89 & 83.40  & 58.41  \\
Qwen3-VL-4B                     & 21.02 & 49.98 & 65.37 & 79.98 & 90.72  & 61.41 & 33.82 & 46.45 & 57.76 & 72.53 & 83.44  & 58.80  \\
Qwen3-VL-8B                     & 20.99 & 50.18 & 65.33 & 80.28 & 90.66  & 61.49 & 33.71 & 46.45 & 57.80 & 73.06 & 84.37  & 59.08  \\
\bottomrule
\end{tabular}}
\end{table}

Table~\ref{tab:appendix_scaling_law} reports the detailed geolocalization accuracy for each backbone model configuration on both IM2GPS3K and YFCC4K, complementing the mean accuracy summary presented in the main text. The results are consistent with the trends described in the main text. Within the Qwen2-VL family, scaling from 2B to 7B parameters yields improvements across all distance thresholds on both datasets. Within the Qwen3-VL family, scaling from 2B to 8B produces steady improvements, and the gains are more evenly distributed across thresholds compared to Qwen2-VL family, suggesting that the architectural improvements in Qwen3-VL lead to more balanced routing capability across different spatial granularities. Notably, even the smallest configuration, Qwen2-VL-2B, consistently outperforms the strongest single-paradigm baseline GeoRanker, which further confirms that the routing framework is effective regardless of backbone capacity.

\section{Data Efficiency Analysis with All Geographic Levels} \label{sec:appendix_data_efficiency}

\begin{figure}
    \centering
    \includegraphics[width=\linewidth]{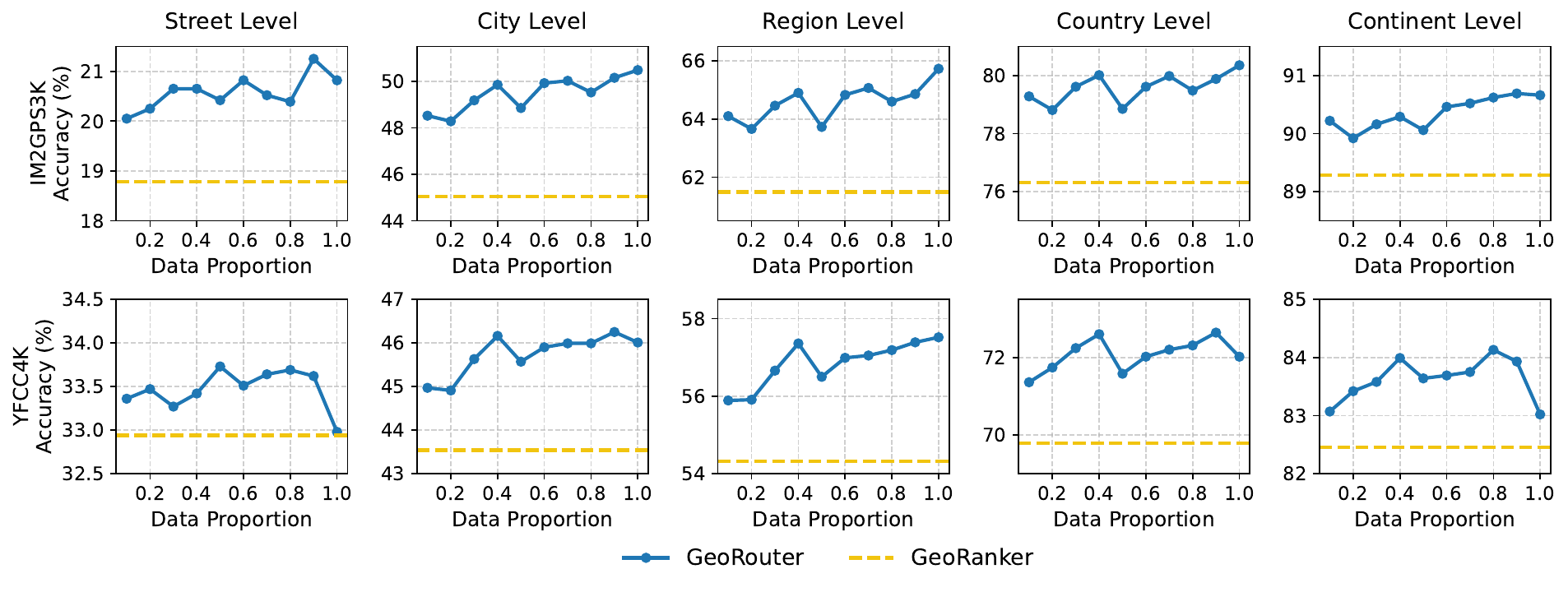}
    \caption{Detailed data efficiency analysis across five geographic levels on the IM2GPS3K (top row) and YFCC4K (bottom row) datasets. The solid blue line represents the accuracy of GeoRouter trained on varying proportions of the dataset, while the dashed yellow line indicates the performance of the state-of-the-art baseline GeoRanker.}
    \label{fig:appendix_data_efficiency}
\end{figure}

In the main text, we present the mean geolocalization accuracy of GeoRouter under varying volumes of training data. Figure~\ref{fig:appendix_data_efficiency} provides a comprehensive breakdown of these data efficiency results across all five geographic levels for both the IM2GPS3K and YFCC4K datasets. The detailed evaluation yields two consistent observations that align with the aggregated results. First, the advantage of GeoRouter over the state-of-the-art baseline, GeoRanker, is robust across all spatial scales. Even when trained on only 10\% of the available dataset, GeoRouter outperforms GeoRanker at every individual distance threshold on both datasets. Second, as the proportion of training data increases, the geolocalization accuracy generally improves across the various spatial levels. While minor fluctuations occur at specific granularities (for example, the Street level on the YFCC4K dataset), the overall positive trend confirms that the dynamic routing mechanism stably benefits from increased geographic diversity in the training data.

\section{Inference Overhead Analysis} \label{sec:appendix_time_efficiency}

\begin{table}
\centering
\caption{Inference overhead of GeoRouter on the IM2GPS3K and YFCC4K datasets.}
\label{tab:appendix_inference_overhead}
\resizebox{\linewidth}{!}{
\begin{tabular}{cccccc} 
\toprule
\multirow{2}{*}{GPU} & \multirow{2}{*}{Batch Size} & \multicolumn{2}{c}{IM2GPS3K}                & \multicolumn{2}{c}{YFCC4K}                   \\ 
\cmidrule{3-6}
                     &                             & GPU Memory Consumption & Inference Speed    & GPU Memory Consumption & Inference Speed     \\ 
\midrule
NVIDIA H800 $\times$ 1        & 16                          & Avg 65 GiB             & 0.1054s per figure & Avg 65 GiB             & 0.0551s per figure  \\
\bottomrule
\end{tabular}}
\end{table}

In this section, we evaluate the computational overhead introduced by the routing component during the inference stage. Table~\ref{tab:appendix_inference_overhead} details the GPU memory consumption and inference speed of GeoRouter. The evaluation is conducted on a single NVIDIA H800 GPU using a batch size of 16.

%% file: 0Abstract.bbl
\begin{thebibliography}{10}

\bibitem{wilson2021visual}
Daniel Wilson, Xiaohan Zhang, Waqas Sultani, and Safwan Wshah.
\newblock Visual and object geo-localization: A comprehensive survey.
\newblock {\em arXiv preprint arXiv:2112.15202}, 2021.

\bibitem{li2025pixels}
Lingyao Li, Runlong Yu, Qikai Hu, Bowei Li, Min Deng, Yang Zhou, and Xiaowei Jia.
\newblock From pixels to places: A systematic benchmark for evaluating image geolocalization ability in large language models.
\newblock {\em arXiv preprint arXiv:2508.01608}, 2025.

\bibitem{noh2017large}
Hyeonwoo Noh, Andre Araujo, Jack Sim, Tobias Weyand, and Bohyung Han.
\newblock Large-scale image retrieval with attentive deep local features.
\newblock In {\em Proceedings of the IEEE international conference on computer vision}, pages 3456--3465, 2017.

\bibitem{cao2020unifying}
Bingyi Cao, Andre Araujo, and Jack Sim.
\newblock Unifying deep local and global features for image search.
\newblock In {\em European conference on computer vision}, pages 726--743. Springer, 2020.

\bibitem{tan2021instance}
Fuwen Tan, Jiangbo Yuan, and Vicente Ordonez.
\newblock Instance-level image retrieval using reranking transformers.
\newblock In {\em proceedings of the IEEE/CVF international conference on computer vision}, pages 12105--12115, 2021.

\bibitem{lee2022correlation}
Seongwon Lee, Hongje Seong, Suhyeon Lee, and Euntai Kim.
\newblock Correlation verification for image retrieval.
\newblock In {\em Proceedings of the IEEE/CVF conference on computer vision and pattern recognition}, pages 5374--5384, 2022.

\bibitem{shao2023global}
Shihao Shao, Kaifeng Chen, Arjun Karpur, Qinghua Cui, Andr{\'e} Araujo, and Bingyi Cao.
\newblock Global features are all you need for image retrieval and reranking.
\newblock In {\em Proceedings of the IEEE/CVF International Conference on Computer Vision}, pages 11036--11046, 2023.

\bibitem{liu2024image}
Yi~Liu, Junchen Ding, Gelei Deng, Yuekang Li, Tianwei Zhang, Weisong Sun, Yaowen Zheng, Jingquan Ge, and Yang Liu.
\newblock Image-based geolocation using large vision-language models.
\newblock {\em arXiv preprint arXiv:2408.09474}, 2024.

\bibitem{li2025cross}
Hao Li, Fabian Deuser, Wenping Yin, Xuanshu Luo, Paul Walther, Gengchen Mai, Wei Huang, and Martin Werner.
\newblock Cross-view geolocalization and disaster mapping with street-view and vhr satellite imagery: A case study of hurricane ian.
\newblock {\em ISPRS Journal of Photogrammetry and Remote Sensing}, 220:841--854, 2025.

\bibitem{kadha2025unravelling}
VijayaKumar Kadha, Sambit Bakshi, and Santos~Kumar Das.
\newblock Unravelling digital forgeries: A systematic survey on image manipulation detection and localization.
\newblock {\em ACM Computing Surveys}, 57(12):1--36, 2025.

\bibitem{wang2024llmgeo}
Zhiqiang Wang, Dejia Xu, Rana Muhammad~Shahroz Khan, Yanbin Lin, Zhiwen Fan, and Xingquan Zhu.
\newblock Llmgeo: Benchmarking large language models on image geolocation in-the-wild.
\newblock {\em arXiv preprint arXiv:2405.20363}, 2024.

\bibitem{lin2013cross}
Tsung-Yi Lin, Serge Belongie, and James Hays.
\newblock Cross-view image geolocalization.
\newblock In {\em Proceedings of the IEEE Conference on Computer Vision and Pattern Recognition}, pages 891--898, 2013.

\bibitem{astruc2024openstreetview}
Guillaume Astruc, Nicolas Dufour, Ioannis Siglidis, Constantin Aronssohn, Nacim Bouia, Stephanie Fu, Romain Loiseau, Van~Nguyen Nguyen, Charles Raude, Elliot Vincent, et~al.
\newblock Openstreetview-5m: The many roads to global visual geolocation.
\newblock In {\em Proceedings of the IEEE/CVF Conference on Computer Vision and Pattern Recognition}, pages 21967--21977, 2024.

\bibitem{jay2025evaluating}
Neel Jay, Hieu~Minh Nguyen, Trung~Dung Hoang, and Jacob Haimes.
\newblock Evaluating precise geolocation inference capabilities of vision language models.
\newblock {\em arXiv preprint arXiv:2502.14412}, 2025.

\bibitem{pandegeochain}
Nilay Pande, Sahiti Yerramilli, Jayant~Sravan Tamarapalli, and Rynaa Grover.
\newblock Geochain: Multimodal chain-of-thought for geographic reasoning.
\newblock In {\em The First Workshop on Multimodal Knowledge and Language Modeling}.

\bibitem{talreja2026georc}
Mohit Talreja, Joshua Diao, Jim~Thannikary James, Radu Casapu, Tejas Santanam, Ethan Mendes, Alan Ritter, Wei Xu, and James Hays.
\newblock Georc: A benchmark for geolocation reasoning chains.
\newblock {\em arXiv preprint arXiv:2601.21278}, 2026.

\bibitem{haas2023learning}
Lukas Haas, Silas Alberti, and Michal Skreta.
\newblock Learning generalized zero-shot learners for open-domain image geolocalization.
\newblock {\em arXiv preprint arXiv:2302.00275}, 2023.

\bibitem{vivanco2023geoclip}
Vicente Vivanco~Cepeda, Gaurav~Kumar Nayak, and Mubarak Shah.
\newblock Geoclip: Clip-inspired alignment between locations and images for effective worldwide geo-localization.
\newblock {\em Advances in Neural Information Processing Systems}, 36:8690--8701, 2023.

\bibitem{jia2025georanker}
Pengyue Jia, Seongheon Park, Song Gao, Xiangyu Zhao, and Sharon Li.
\newblock Georanker: Distance-aware ranking for worldwide image geolocalization.
\newblock {\em arXiv preprint arXiv:2505.13731}, 2025.

\bibitem{fang2026geomr}
Jian Fang, Siyi Qian, and Shaohui Liu.
\newblock Geomr: Integrating image geographic features and human reasoning knowledge for image geolocalization.
\newblock {\em Knowledge-Based Systems}, page 115391, 2026.

\bibitem{zhou2024img2loc}
Zhongliang Zhou, Jielu Zhang, Zihan Guan, Mengxuan Hu, Ni~Lao, Lan Mu, Sheng Li, and Gengchen Mai.
\newblock Img2loc: Revisiting image geolocalization using multi-modality foundation models and image-based retrieval-augmented generation.
\newblock In {\em Proceedings of the 47th international acm sigir conference on research and development in information retrieval}, pages 2749--2754, 2024.

\bibitem{jia2024g3}
Pengyue Jia, Yiding Liu, Xiaopeng Li, Xiangyu Zhao, Yuhao Wang, Yantong Du, Xiao Han, Xuetao Wei, Shuaiqiang Wang, and Dawei Yin.
\newblock G3: an effective and adaptive framework for worldwide geolocalization using large multi-modality models.
\newblock {\em Advances in Neural Information Processing Systems}, 37:53198--53221, 2024.

\bibitem{ghasemi2025geotoken}
Narges Ghasemi, Amir Ziashahabi, Salman Avestimehr, and Cyrus Shahabi.
\newblock Geotoken: Hierarchical geolocalization of images via next token prediction.
\newblock {\em arXiv preprint arXiv:2511.01082}, 2025.

\bibitem{wu2026vision}
Biao Wu, Meng Fang, Ling Chen, Ke~Xu, Tao Cheng, and Jun Wang.
\newblock Vision-language reasoning for geolocalization: A reinforcement learning approach.
\newblock {\em arXiv preprint arXiv:2601.00388}, 2026.

\bibitem{yu2026locatability}
Bo~Yu, Fengze Yang, Yiming Liu, Chao Wang, Xuewen Luo, Taozhe Li, Ruimin Ke, Xiaofan Zhou, and Chenxi Liu.
\newblock Locatability-guided adaptive reasoning for image geo-localization with vision-language models.
\newblock {\em arXiv preprint arXiv:2603.13628}, 2026.

\bibitem{ji2026thinking}
Yuxiang Ji, Yong Wang, Ziyu Ma, Yiming Hu, Hailang Huang, Xuecai Hu, Guanhua Chen, Liaoni Wu, and Xiangxiang Chu.
\newblock Thinking with map: Reinforced parallel map-augmented agent for geolocalization.
\newblock {\em arXiv preprint arXiv:2601.05432}, 2026.

\bibitem{zheng2026learningwanderimprovingglobal}
Yushuo Zheng, Huiyu Duan, Zicheng Zhang, Xiaohong Liu, and Xiongkuo Min.
\newblock Learning to wander: Improving the global image geolocation ability of lmms via actionable reasoning, 2026.

\bibitem{li2024georeasoner}
Ling Li, Yu~Ye, Yao Zhou, Bingchuan Jiang, and Wei Zeng.
\newblock Georeasoner: Geo-localization with reasoning in street views using a large vision-language model.
\newblock {\em arXiv preprint arXiv:2406.18572}, 2024.

\bibitem{li2025recognition}
Ling Li, Yao Zhou, Yuxuan Liang, Fugee Tsung, and Jiaheng Wei.
\newblock Recognition through reasoning: Reinforcing image geo-localization with large vision-language models.
\newblock {\em arXiv preprint arXiv:2506.14674}, 2025.

\bibitem{wang2025gre}
Chun Wang, Xiaojun Ye, Xiaoran Pan, Zihao Pan, Haofan Wang, and Yiren Song.
\newblock Gre suite: Geo-localization inference via fine-tuned vision-language models and enhanced reasoning chains.
\newblock {\em arXiv preprint arXiv:2505.18700}, 2025.

\bibitem{jin2026geoagentlearninggeolocatereinforced}
Modi Jin, Yiming Zhang, Boyuan Sun, Dingwen Zhang, MingMing Cheng, and Qibin Hou.
\newblock Geoagent: Learning to geolocate everywhere with reinforced geographic characteristics, 2026.

\bibitem{shi2026geobayes}
Weimin Shi, Xiang Li, Kaige Li, Junhao Fang, Qiang Zhou, Qichuan Geng, and Zhong Zhou.
\newblock Geobayes: Probabilistic image geo-localization inference via sequential bayesian updating.
\newblock In {\em Proceedings of the AAAI Conference on Artificial Intelligence}, volume~40, pages 8997--9005, 2026.

\bibitem{su2026interpretable}
Yiyang Su and Xiaoming Liu.
\newblock Interpretable perception and reasoning for audiovisual geolocation.
\newblock {\em arXiv preprint arXiv:2603.05708}, 2026.

\bibitem{DBLP:conf/mediaeval/ChoiHLT16}
Jaeyoung Choi, Claudia Hauff, Olivier~Van Laere, and Bart Thomee.
\newblock The placing task at mediaeval 2016.
\newblock In Guillaume Gravier, Claire{-}H{\'{e}}l{\`{e}}ne Demarty, Herv{\'{e}} Bredin, Bogdan Ionescu, Christina Boididou, Emmanuel Dellandr{\'{e}}a, Jaeyoung Choi, Michael Riegler, Richard F.~E. Sutcliffe, Igor Sz{\"{o}}ke, Gareth J.~F. Jones, and Martha~A. Larson, editors, {\em Working Notes Proceedings of the MediaEval 2016 Workshop, Hilversum, The Netherlands, October 20-21, 2016}, volume 1739 of {\em {CEUR} Workshop Proceedings}. CEUR-WS.org, 2016.

\bibitem{comanici2025gemini}
Gheorghe Comanici, Eric Bieber, Mike Schaekermann, Ice Pasupat, Noveen Sachdeva, Inderjit Dhillon, Marcel Blistein, Ori Ram, Dan Zhang, Evan Rosen, et~al.
\newblock Gemini 2.5: Pushing the frontier with advanced reasoning, multimodality, long context, and next generation agentic capabilities.
\newblock {\em arXiv preprint arXiv:2507.06261}, 2025.

\bibitem{dufour2025around}
Nicolas Dufour, Vicky Kalogeiton, David Picard, and Loic Landrieu.
\newblock Around the world in 80 timesteps: A generative approach to global visual geolocation.
\newblock In {\em Proceedings of the Computer Vision and Pattern Recognition Conference}, pages 23016--23026, 2025.

\bibitem{wanglocdiff}
Zhangyu Wang, Zeping Liu, Jielu Zhang, Zhongliang Zhou, Qian Cao, Nemin Wu, Lan Mu, Yang Song, Yiqun Xie, Ni~Lao, et~al.
\newblock Locdiff: Identifying locations on earth by diffusing in the hilbert space.
\newblock In {\em The Thirty-ninth Annual Conference on Neural Information Processing Systems}.

\bibitem{haas2024pigeon}
Lukas Haas, Michal Skreta, Silas Alberti, and Chelsea Finn.
\newblock Pigeon: Predicting image geolocations.
\newblock In {\em Proceedings of the IEEE/CVF Conference on Computer Vision and Pattern Recognition}, pages 12893--12902, 2024.

\bibitem{dou2024gaga}
Zhiyang Dou, Zipeng Wang, Xumeng Han, Guorong Li, Zhipei Huang, and Zhenjun Han.
\newblock Gaga: Towards interactive global geolocation assistant.
\newblock {\em arXiv preprint arXiv:2412.08907}, 2024.

\bibitem{hays2008im2gps}
James Hays and Alexei~A Efros.
\newblock Im2gps: estimating geographic information from a single image.
\newblock In {\em 2008 ieee conference on computer vision and pattern recognition}, pages 1--8. IEEE, 2008.

\bibitem{thomee2016yfcc100m}
Bart Thomee, David~A Shamma, Gerald Friedland, Benjamin Elizalde, Karl Ni, Douglas Poland, Damian Borth, and Li-Jia Li.
\newblock Yfcc100m: The new data in multimedia research.
\newblock {\em Communications of the ACM}, 59(2):64--73, 2016.

\bibitem{sarkar2024gomaa}
Anindya Sarkar, Srikumar Sastry, Aleksis Pirinen, Chongjie Zhang, Nathan Jacobs, and Yevgeniy Vorobeychik.
\newblock Gomaa-geo: Goal modality agnostic active geo-localization.
\newblock {\em Advances in Neural Information Processing Systems}, 37:104934--104964, 2024.

\bibitem{vo2017revisiting}
Nam Vo, Nathan Jacobs, and James Hays.
\newblock Revisiting im2gps in the deep learning era.
\newblock In {\em Proceedings of the IEEE international conference on computer vision}, pages 2621--2630, 2017.

\bibitem{voulodimos2018deep}
Athanasios Voulodimos, Nikolaos Doulamis, Anastasios Doulamis, and Eftychios Protopapadakis.
\newblock Deep learning for computer vision: A brief review.
\newblock {\em Computational intelligence and neuroscience}, 2018(1):7068349, 2018.

\bibitem{he2016deep}
Kaiming He, Xiangyu Zhang, Shaoqing Ren, and Jian Sun.
\newblock Deep residual learning for image recognition.
\newblock In {\em Proceedings of the IEEE conference on computer vision and pattern recognition}, pages 770--778, 2016.

\bibitem{mai2024opportunities}
Gengchen Mai, Weiming Huang, Jin Sun, Suhang Song, Deepak Mishra, Ninghao Liu, Song Gao, Tianming Liu, Gao Cong, Yingjie Hu, et~al.
\newblock On the opportunities and challenges of foundation models for geoai (vision paper).
\newblock {\em ACM Transactions on Spatial Algorithms and Systems}, 10(2):1--46, 2024.

\bibitem{janowicz2020geoai}
Krzysztof Janowicz, Song Gao, Grant McKenzie, Yingjie Hu, and Budhendra Bhaduri.
\newblock Geoai: spatially explicit artificial intelligence techniques for geographic knowledge discovery and beyond, 2020.

\bibitem{weyand2016planet}
Tobias Weyand, Ilya Kostrikov, and James Philbin.
\newblock Planet-photo geolocation with convolutional neural networks.
\newblock In {\em European conference on computer vision}, pages 37--55. Springer, 2016.

\bibitem{seo2018cplanet}
Paul~Hongsuck Seo, Tobias Weyand, Jack Sim, and Bohyung Han.
\newblock Cplanet: Enhancing image geolocalization by combinatorial partitioning of maps.
\newblock In {\em Proceedings of the European Conference on Computer Vision (ECCV)}, pages 536--551, 2018.

\bibitem{muller2018geolocation}
Eric Muller-Budack, Kader Pustu-Iren, and Ralph Ewerth.
\newblock Geolocation estimation of photos using a hierarchical model and scene classification.
\newblock In {\em Proceedings of the European conference on computer vision (ECCV)}, pages 563--579, 2018.

\bibitem{pramanick2022world}
Shraman Pramanick, Ewa~M Nowara, Joshua Gleason, Carlos~D Castillo, and Rama Chellappa.
\newblock Where in the world is this image? transformer-based geo-localization in the wild.
\newblock In {\em European Conference on Computer Vision}, pages 196--215. Springer, 2022.

\bibitem{clark2023we}
Brandon Clark, Alec Kerrigan, Parth~Parag Kulkarni, Vicente~Vivanco Cepeda, and Mubarak Shah.
\newblock Where we are and what we're looking at: Query based worldwide image geo-localization using hierarchies and scenes.
\newblock In {\em Proceedings of the IEEE/CVF Conference on Computer Vision and Pattern Recognition}, pages 23182--23190, 2023.

\bibitem{workman2015wide}
Scott Workman, Richard Souvenir, and Nathan Jacobs.
\newblock Wide-area image geolocalization with aerial reference imagery.
\newblock In {\em Proceedings of the IEEE International Conference on Computer Vision}, pages 3961--3969, 2015.

\bibitem{liu2019lending}
Liu Liu and Hongdong Li.
\newblock Lending orientation to neural networks for cross-view geo-localization.
\newblock In {\em Proceedings of the IEEE/CVF conference on computer vision and pattern recognition}, pages 5624--5633, 2019.

\bibitem{zhu2021vigor}
Sijie Zhu, Taojiannan Yang, and Chen Chen.
\newblock Vigor: Cross-view image geo-localization beyond one-to-one retrieval.
\newblock In {\em Proceedings of the IEEE/CVF Conference on Computer Vision and Pattern Recognition}, pages 3640--3649, 2021.

\bibitem{lin2022joint}
Jinliang Lin, Zhedong Zheng, Zhun Zhong, Zhiming Luo, Shaozi Li, Yi~Yang, and Nicu Sebe.
\newblock Joint representation learning and keypoint detection for cross-view geo-localization.
\newblock {\em IEEE Transactions on Image Processing}, 31:3780--3792, 2022.

\bibitem{zhu2022transgeo}
Sijie Zhu, Mubarak Shah, and Chen Chen.
\newblock Transgeo: Transformer is all you need for cross-view image geo-localization.
\newblock In {\em Proceedings of the IEEE/CVF Conference on Computer Vision and Pattern Recognition}, pages 1162--1171, 2022.

\bibitem{zhang2023cross}
Xiaohan Zhang, Xingyu Li, Waqas Sultani, Yi~Zhou, and Safwan Wshah.
\newblock Cross-view geo-localization via learning disentangled geometric layout correspondence.
\newblock In {\em Proceedings of the AAAI conference on artificial intelligence}, volume~37, pages 3480--3488, 2023.

\bibitem{ye2025cross}
Junyan Ye, Honglin Lin, Leyan Ou, Dairong Chen, Zihao Wang, Qi~Zhu, Conghui He, and Weijia Li.
\newblock Where am i? cross-view geo-localization with natural language descriptions.
\newblock In {\em Proceedings of the IEEE/CVF International Conference on Computer Vision}, pages 5890--5900, 2025.

\bibitem{tian2017cross}
Yicong Tian, Chen Chen, and Mubarak Shah.
\newblock Cross-view image matching for geo-localization in urban environments.
\newblock In {\em Proceedings of the IEEE Conference on Computer Vision and Pattern Recognition}, pages 3608--3616, 2017.

\bibitem{shi2020looking}
Yujiao Shi, Xin Yu, Dylan Campbell, and Hongdong Li.
\newblock Where am i looking at? joint location and orientation estimation by cross-view matching.
\newblock In {\em Proceedings of the IEEE/CVF Conference on Computer Vision and Pattern Recognition}, pages 4064--4072, 2020.

\bibitem{zhu2023r2former}
Sijie Zhu, Linjie Yang, Chen Chen, Mubarak Shah, Xiaohui Shen, and Heng Wang.
\newblock R2former: Unified retrieval and reranking transformer for place recognition.
\newblock In {\em Proceedings of the IEEE/CVF conference on computer vision and pattern recognition}, pages 19370--19380, 2023.

\bibitem{jia2026spotagent}
Furong Jia, Ling Dai, Wenjin Deng, Fan Zhang, Chen Hu, Daxin Jiang, and Yu~Liu.
\newblock Spotagent: Grounding visual geo-localization in large vision-language models through agentic reasoning.
\newblock {\em arXiv preprint arXiv:2602.09463}, 2026.

\bibitem{kotthoff2016algorithm}
Lars Kotthoff.
\newblock Algorithm selection for combinatorial search problems: A survey.
\newblock In {\em Data mining and constraint programming: Foundations of a cross-disciplinary approach}, pages 149--190. Springer, 2016.

\bibitem{kerschke2019automated}
Pascal Kerschke, Holger~H Hoos, Frank Neumann, and Heike Trautmann.
\newblock Automated algorithm selection: Survey and perspectives.
\newblock {\em Evolutionary computation}, 27(1):3--45, 2019.

\bibitem{varangot2025doing}
Clovis Varangot-Reille, Christophe Bouvard, Antoine Gourru, Mathieu Ciancone, Marion Schaeffer, and Fran{\c{c}}ois Jacquenet.
\newblock Doing more with less: A survey on routing strategies for resource optimisation in large language model-based systems.
\newblock {\em arXiv preprint arXiv:2502.00409}, 2025.

\bibitem{jang2023exploring}
Joel Jang, Seungone Kim, Seonghyeon Ye, Doyoung Kim, Lajanugen Logeswaran, Moontae Lee, Kyungjae Lee, and Minjoon Seo.
\newblock Exploring the benefits of training expert language models over instruction tuning.
\newblock In {\em International Conference on Machine Learning}, pages 14702--14729. PMLR, 2023.

\bibitem{ong2024routellm}
Isaac Ong, Amjad Almahairi, Vincent Wu, Wei-Lin Chiang, Tianhao Wu, Joseph~E Gonzalez, M~Waleed Kadous, and Ion Stoica.
\newblock Routellm: Learning to route llms with preference data.
\newblock {\em arXiv preprint arXiv:2406.18665}, 2024.

\bibitem{zhuang2024embedllm}
Richard Zhuang, Tianhao Wu, Zhaojin Wen, Andrew Li, Jiantao Jiao, and Kannan Ramchandran.
\newblock Embedllm: Learning compact representations of large language models.
\newblock {\em arXiv preprint arXiv:2410.02223}, 2024.

\bibitem{ding2024hybrid}
Dujian Ding, Ankur Mallick, Chi Wang, Robert Sim, Subhabrata Mukherjee, Victor Ruhle, Laks~VS Lakshmanan, and Ahmed~Hassan Awadallah.
\newblock Hybrid llm: Cost-efficient and quality-aware query routing.
\newblock {\em arXiv preprint arXiv:2404.14618}, 2024.

\bibitem{hu2024routerbench}
Qitian~Jason Hu, Jacob Bieker, Xiuyu Li, Nan Jiang, Benjamin Keigwin, Gaurav Ranganath, Kurt Keutzer, and Shriyash~Kaustubh Upadhyay.
\newblock Routerbench: A benchmark for multi-llm routing system.
\newblock {\em arXiv preprint arXiv:2403.12031}, 2024.

\bibitem{sikeridis2025pickllm}
Dimitrios Sikeridis, Dennis Ramdass, and Pranay Pareek.
\newblock Pickllm: Context-aware rl-assisted large language model routing.
\newblock In {\em International Workshop on AI for Transportation}, pages 227--239. Springer, 2025.

\bibitem{lu2024routing}
Keming Lu, Hongyi Yuan, Runji Lin, Junyang Lin, Zheng Yuan, Chang Zhou, and Jingren Zhou.
\newblock Routing to the expert: Efficient reward-guided ensemble of large language models.
\newblock In {\em Proceedings of the 2024 Conference of the North American Chapter of the Association for Computational Linguistics: Human Language Technologies (Volume 1: Long Papers)}, pages 1964--1974, 2024.

\bibitem{li2024retrieval}
Zhuowan Li, Cheng Li, Mingyang Zhang, Qiaozhu Mei, and Michael Bendersky.
\newblock Retrieval augmented generation or long-context llms? a comprehensive study and hybrid approach.
\newblock {\em arXiv preprint arXiv:2407.16833}, 2024.

\bibitem{patil2024gorilla}
Shishir~G Patil, Tianjun Zhang, Xin Wang, and Joseph~E Gonzalez.
\newblock Gorilla: Large language model connected with massive apis.
\newblock {\em Advances in Neural Information Processing Systems}, 37:126544--126565, 2024.

\bibitem{hu2022lora}
Edward~J Hu, Yelong Shen, Phillip Wallis, Zeyuan Allen-Zhu, Yuanzhi Li, Shean Wang, Liang Wang, Weizhu Chen, et~al.
\newblock Lora: Low-rank adaptation of large language models.
\newblock {\em Iclr}, 1(2):3, 2022.

\bibitem{jia2025geoarena}
Pengyue Jia, Yingyi Zhang, Xiangyu Zhao, and Sharon Li.
\newblock Geoarena: An open platform for benchmarking large vision-language models on worldwide image geolocalization.
\newblock {\em arXiv preprint arXiv:2509.04334}, 2025.

\bibitem{loshchilov2017decoupled}
Ilya Loshchilov and Frank Hutter.
\newblock Decoupled weight decay regularization.
\newblock {\em arXiv preprint arXiv:1711.05101}, 2017.

\end{thebibliography}
